\documentclass[11pt]{article} % add ,darft for fradft verison

\usepackage[final]{acl} % change to review before submission

\usepackage[T1]{fontenc}
\usepackage[utf8]{inputenc}
\usepackage[babel=true, expansion=true, kerning=true, tracking=true,spacing=true, verbose=silent]{microtype}                     
\usepackage{inconsolata}
\usepackage{graphicx}
\usepackage{times}
\usepackage{latexsym}

\usepackage{custom} 
\usepackage{rotating}

\definecolor{Case1Color}{HTML}{FFA52F}  % Orange
\definecolor{Case2Color}{HTML}{24A7A1}  % Teal
\definecolor{Case3Color}{HTML}{562B35}  % Dark maroon
\definecolor{Case4Color}{HTML}{FF7ED1}  % Pink

\title{Knowing the Facts but Choosing the Shortcut:\\ Understanding How Large Language Models Compare Entities}

\author{
  Hans Hergen Lehmann$^{1}$, 
  Jae Hee Lee$^{1}$\thanks{Corresponding author}, 
  Steven Schockaert$^{2}$, 
  Stefan Wermter$^{1}$\\
  $^{1}$University of Hamburg \quad
  $^{2}$Cardiff University\\
}

\begin{document}
\maketitle

\begin{abstract}
    Large Language Models (LLMs) are increasingly used for knowledge-based reasoning tasks, yet understanding when they rely on genuine knowledge versus superficial heuristics remains challenging. We investigate this question
    %through entity comparison tasks 
    by asking models to compare entities along numerical attributes (e.g., ``Which river is longer, the Danube or the Nile?''), which offers clear ground truth for systematic analysis. Despite having sufficient numerical knowledge to answer correctly, LLMs frequently make predictions that contradict this knowledge. We identify three heuristic biases that strongly influence model predictions: entity popularity, mention order, and semantic co-occurrence. For smaller models, a simple logistic regression using only these surface cues predicts model choices more accurately than the model's own numerical predictions, suggesting heuristics largely override principled reasoning. Crucially, we find that larger models (32B parameters) selectively rely on numerical knowledge when it is more reliable, while smaller models (7--8B parameters) show no such discrimination, which explains why larger models outperform smaller ones even when the smaller models possess more accurate knowledge. Chain-of-thought prompting steers all models towards using the numerical features across all model sizes.
\end{abstract}

\section{Introduction}

There is an ongoing debate about the extent to which LLMs \emph{understand} language, and the world more generally \cite{mitchell2023debate, ray_choudhury_machine_2022}. Two contrasting views have been put forward: the \emph{world-model} view holds that LLMs internalize structured knowledge about the world, which they can deploy when prompted \cite{DBLP:conf/iclr/0002HBVPW23, DBLP:conf/icml/JinR24}, while the \emph{statistical-parrot} view argues that outputs are largely driven by surface cues \cite{bender_dangers_2021, DBLP:conf/er/Saba23}. While it seems reasonable to assume that the truth is somewhere in the middle, untangling \emph{when} LLMs rely on surface cues and when they rely on genuine understanding is often hard. In this paper, we therefore focus on a simple controlled setting, within which this question can be studied more systematically, namely the problem of comparing entities along some numerical attribute (e.g., ``which country has the highest population, France or Germany?'').

This problem setting has several key advantages. First, for the attributes that we consider, there is a unique and objective ground truth. Second, as we can randomly sample entity pairs from a large set of candidates, we can straightforwardly construct test sets that are balanced and orthogonal (i.e., where the presence of one feature is independent of the presence of another feature), which is important for systematic analysis. Moreover, as most entity pairs are not directly compared anywhere on the Web, the impact of pure memorization on the performance of the model should be negligible. Finally, the required world knowledge and associated reasoning process are clear and simple. This means, for instance, that we can straightforwardly identify cases where the LLM has the required knowledge (i.e., knows the correct numerical values), and thus we can distinguish errors due to a lack of knowledge from erroneous reasoning.

We start our analysis by asking: \hyperref[sec:gener-model-perf]{\emph{do LLMs use numerical attributes for pairwise comparisons?}} (see \Cref{sec:gener-model-perf}). We show that the pairwise predictions are often inconsistent with predicted attribute values, which suggests that LLMs do not consistently exploit their internal knowledge about these attributes. This is despite the fact that using the predicted attribute values would lead to more accurate results. We also note that the accuracy of pairwise predictions improves as model sizes are increased, but the same is not always true when it comes to the accuracy of predicted numerical attribute values. In other words, larger models perform better (as could be expected), but this is \emph{not} due to having more accurate knowledge.

To better understand the underlying reasons, we ask our next question: \hyperref[sec:biases]{\emph{how susceptible are LLMs to heuristic biases when answering pairwise comparison queries?}} (see \Cref{sec:biases}). We show that pairwise predictions are strongly biased by three types of surface cues: the \emph{position} of an entity in the prompt, entity \emph{popularity}, and shallow \emph{co-occurrence statistics}.
Since these heuristics often align with the ground truth in natural data, we control for
this alignment using a balanced, orthogonal subset.
We then ask: \hyperref[sec:meta-predictor]{\emph{to what extent can the pairwise predictions be \emph{explained} by these surface cues?} (see \Cref{sec:meta-predictor}}). We find that the vast majority of model predictions can either be explained by the predicted numerical features or by the above three types of surface cues. This suggests that the considered models sometimes compare numerical values while at other times falling back on surface cues. We find that larger models are more likely to rely on the numerical values when these values are more accurate (i.e., closer to the ground truth numerical values), whereas no such effect was observed for the smallest models (i.e., they rely on shortcuts even if they know the numerical values). This difference explains why the largest models outperform smaller models on pairwise predictions, even though they do not always outperform smaller models in terms of predicting the numerical attribute values.

Finally, we ask: \hyperref[sec:thinking]{\emph{does chain-of-thought based reasoning help models use their own numerical predictions more faithfully when making pairwise judgments?}} (see \Cref{sec:thinking}). We find that allowing a model to verbalize its reasoning process indeed leads to a more consistent use of numerical attributes, which narrows the performance gap between models of different sizes.

Our findings reveal that the superior performance of larger models stems not from more accurate knowledge, but from their ability to strategically choose when to rely on that knowledge versus when to fall back on heuristics. This suggests that scaling improvements in LLMs may be driven as much by better strategy selection
as by knowledge acquisition itself, with important implications for understanding and improving model reliability.

\section{Experimental Setup}
We focus on an entity comparison task, where we prompt an LLM with pairwise comparison questions (e.g., ``Which river is longer, the Danube or the Rhine?'') and evaluate whether the model selected the correct item according to the ground truth.\footnote{Code to reproduce our experiments is available at
%\url{...}
\url{https://github.com/HeLehm/facts-vs-shortcuts}
} 

\paragraph{Datasets.}
To obtain a sufficiently large set of test queries, we collected data on 10 different numerical attributes across diverse entity types from Wikidata\footnote{\url{https://www.wikidata.org}}. The selected attributes cover a range of domains, such as geography (e.g., river length, population of countries and cities) and science (e.g., atomic numbers), which are listed in \cref{tab:dataset_info}. For each attribute, we sampled up to 1000 entities. For each attribute, we begin by selecting the most popular entities, based on their QRank\footnote{QRank is a popularity ranking for Wikidata entities computed by aggregating page view statistics. See \url{https://qrank.toolforge.org}.} score.  To obtain a set of entity pairs that span a range of difficulty levels, we employ a stratified sampling approach. Specifically, we first sort all entities by their ground-truth attribute values and divide them into two equal-sized bins: a lower-value bin and a higher-value bin. For every entity in our sample, we construct two comparison pairs by randomly selecting one partner from each bin. This ensures our dataset includes both challenging near-tie comparisons and clearer-cut distinctions.

\begin{table}[t]
    \footnotesize
    \centering
      \begin{tabular}{llr}
        \toprule
        \textbf{Dataset} & \textbf{Attribute}   & \textbf{Entities} \\
        \midrule
        Atoms            & Atomic number        & 118                   \\
        Buildings        & Height               & 1000                  \\
        Cities           & Population           & 1000                  \\
        Countries        & Population           & 196                   \\
        Mountains        & Elevation            & 997                   \\
        Peppers          & Scoville heat unit   & 45                    \\
        People           & \# Followers         & 999                   \\
        Rivers           & Length               & 999                   \\
        Stadiums         & Capacity             & 999                   \\
        Universities     & \# Enrolled students & 1000                  \\
        \bottomrule
    \end{tabular}
    \caption{Overview of the considered datasets.}
    \label{tab:dataset_info}
\end{table}

\paragraph{Prompting Strategy.}
The performance of LLMs can be sensitive to the choice of prompt. For this reason, each entity pair is evaluated across six prompt templates. The first three templates ask which of the two entities has the highest attribute value. The remaining three templates ask for the entity with the lowest value. Furthermore, for each template, we prompt the model twice for every entity pair, i.e., once for each of the possible entity orderings (e.g., \texttt{(Danube, Nile)} and \texttt{(Nile, Danube)}). In total, we thus have $6 \times 2 = 12$ prompts per entity pair. We list the full set of prompt templates and describe the strategy for parsing the answers from the model's output in \cref{sec:prompts,sec:parsing_details}. We also analyze the sensitivity of our results to the choice of prompt templates in \cref{sec:prompt-sensitivity}. In addition to prompting for pairwise comparisons, we prompt the model to predict the numerical attribute values of the entities. To this end, we use three numerical extraction templates for each attribute and select the prediction with the lowest perplexity (i.e., we select the model's most confident numerical estimate). We analyze the error of these numerical predictions in \cref{sec:numex_error}.

\paragraph{Evaluation Metrics.}
We assess model performance along three dimensions. First, we measure \emph{pairwise accuracy}, defined as the proportion of pairwise predictions that are correct according to the ground truth. Second, we compute \emph{internal consistency}, which we define as the proportion of pairwise predictions that are in agreement with the ranking implied by the model's own numerical predictions. Finally, we evaluate \emph{numerical accuracy}, which evaluates the quality of the model's predicted attribute values. It is defined as the proportion of pairwise comparisons for which the ranking implied by the predicted numerical values agrees with the ground truth ranking. To ensure comparability, we remove all samples for which the model did not produce a valid answer, either in the pairwise or numerical setting\footnote{\cref{sec:parsing_details} provides a detailed description of how the parsing was done, and an analysis of how often the model did not yield a valid answer.}. As a result, all metrics are computed over the same filtered set of samples.

\paragraph{Models.}
We experiment with models of different families and sizes: Llama3-1B, Llama3-8B \cite{grattafiori_llama_2024}, OLMo2-1B, OLMo2-7B, OLMo2-32B \cite{DBLP:journals/corr/abs-2501-00656}, Qwen3-1.7B, Qwen3-8B, Qwen3-32B \cite{yang2025qwen3technicalreport}, Mistral-7B \cite{DBLP:journals/corr/abs-2310-06825} and Mistral-24B. Full details on these models can be found in \cref{sec:app:model_info}.

\section{Do LLMs Use Numerical Attributes for Pairwise Comparisons?}
\label{sec:gener-model-perf}

\begin{figure}[t]
    \includegraphics[width=\columnwidth]{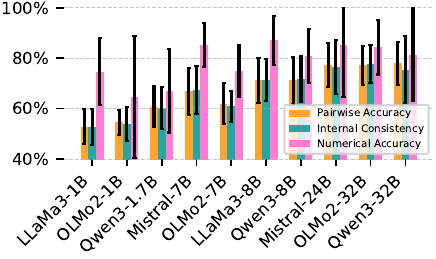}
    \caption{
        Overall performance in terms of pairwise accuracy, internal consistency, and numerical accuracy (mean and standard deviation) 
    }
    \label{fig:general_acc}
\end{figure}

\Cref{fig:general_acc} summarizes the performance of the different language models, averaged across all 10 attributes. A more detailed breakdown can be found in \Cref{sec:app:detailed_acc}. A number of important findings can be observed.
\begin{inparaenum}
    \item Numerical accuracy is consistently and substantially higher than pairwise accuracy, showing that models often make mistakes even when relying on their knowledge of the numerical attributes would produce the correct answer.
    \item For the smallest models, pairwise accuracy is barely above random chance.
    \item Pairwise accuracy increases with model size.
    \item For numerical accuracy, on the other hand, Mistral-7B and Llama3-8B both outperform much bigger models. For these models, the underperformance in terms of pairwise accuracy can thus not be explained by a lack of knowledge (cf.~\Cref{sec:meta-predictor}). This can also be clearly seen from the surprisingly low internal consistency values.
\end{inparaenum}
Overall, the results suggest that LLMs rely on shortcuts when making pairwise predictions, which we further analyze in the next section.

\section{How Susceptible Are Pairwise Predictions to Biases?}
\label{sec:biases}

The previous section showed that LLMs often ignore their own numerical knowledge when ranking entities.
A natural follow-up question is \emph{what, if not the numbers, drives the final choice?}
To investigate, we identify three biases and measure their impact.

\paragraph{Heuristic Cues.}
A first heuristic that LLMs may exploit is that popular entities might have higher values (e.g., cities that are mentioned more often may have higher populations). To analyze this \emph{popularity bias}, we estimate the popularity of each Wikidata entity using its QRank score. We then test whether LLM responses are more accurate when the entity with the highest numerical value is also the most popular one.

Second, LLMs have been found to suffer from \emph{position bias}, favoring responses depending on the order in which they are presented \cite{wang-etal-2024-large-language-models-fair}. We analyze whether a similar bias is also present when comparing entities. To this end, we compare the accuracy across two sets of comparisons: those where the first or second entity has the higher value.

Finally, LLM predictions can be affected by shallow co-occurrence statistics \cite{kang-choi-2023-impact}. To analyze this effect, we rely on the ConceptNet Numberbatch pre-trained word embeddings~\cite{DBLP:conf/aaai/SpeerCH17} as a model of distributional similarity.\footnote{Word embeddings can be seen as a low-rank approximation of co-occurrence statistics and the embedding similarities serve as a convenient proxy for the raw co-occurrence statistics.} For each numerical attribute, we selected 5 adjectives that are indicative of high values (e.g., \emph{longest} for river length) and averaged their embeddings, yielding a vector $\mathbf{v}^+$. We do the same for 5 adjectives that are indicative of low values (e.g., \emph{shortest}) and obtain $\mathbf{v}^-$. We then score entity $e$ as $\cos(\mathbf{e},\mathbf{v}^+-\mathbf{v}^-)$, where $\mathbf{e}$ is the Numberbatch embedding of $e$. Full details of how the scores are obtained can be found in \cref{appendixDetailsCooccurrenceScores}.

\paragraph{Experimental Setup.}

Given the three potential shortcuts, we must design our experiments with care to avoid conflating the model's reliance on surface cues with genuine knowledge of numerical facts. As an example, popularity is often a proxy for magnitude: we remember Mount Everest precisely because it is the tallest peak, and a celebrity's follower count is itself a direct measure of their popularity across social-media platforms. Our data confirm this intuition (see \cref{sec:bias_gt_align} for details). Such correlations mean that surface cues can look like genuine knowledge. Furthermore, we must also consider the possibility that these shortcuts correlate with each other. Mount Everest likely co-occurs with the adjective ``tallest'' in the training data frequently, and it is also more popular than most other mountains. 

Therefore, we need a \emph{balanced} and \emph{orthogonal} design that isolates each cue from the others. \emph{Balance} in this case means that positive and negative cases occur equally frequently (e.g., popularity aligns with the ground truth in exactly half of the comparisons and misaligns in the other half). This ensures that none of the considered heuristic cues has an advantage simply because it happens to be more frequent in the data. \emph{Orthogonality} means that the features vary independently of each other, i.e., all possible combinations of the values of the features appear equally often. Orthogonality helps mitigate aggregation artifacts such as Simpson's paradox, where the apparent effect of a cue might actually be driven by another, correlated factor. If popularity and co-occurrence are correlated, for instance, then the effect we assign to popularity might actually reflect co-occurrence effects, and vice versa. We stress that this protection applies only to the \emph{observed} cues; unmeasured confounders could still induce bias (see \Cref{sec:bias_e_value}).

We construct a \emph{Balanced-Orthogonal Subset} (BOS) as follows. We assign each entity pair with four binary features, which we will refer to as $P$ (popularity), $O$ (order), $C$ (co-occurrence), and $I$ (internal knowledge). We define  \(P=1\) if the entity with the higher ground-truth value is also the more popular one (and $P=0$ otherwise); \(O=1\) if that larger entity appears first in the prompt; \(C=1\) if the entity whose ConceptNet embedding lies closer to the “large” direction is indeed the larger one; \(I=1\) if the ranking implied by the model's extracted numbers matches the ground truth. 
% We study the effect of these features on the pairwise accuracy of the model. 
BOS is constructed by taking the minority count from each of the $2^{4}=16$ \((P,O,C,I)\)-cells within each prompt template and sampling that many instances from every other cell. Information about the size of these subsets can be found in \cref{app:sec:bos_discard}. In BOS, each feature can be toggled independently, assuring that all other features are held constant at a rate of $50\%$ true and $50\%$ false.

To measure the impact of each feature, we adopt the risk ratio (RR)~\cite{rothman_modern_2008}. Let \(Y\) be the ground-truth label and \(\hat{Y}\) the model’s prediction. $F=1$ indicates that a feature is present and $F=0$ means that is not. The RR is defined as follows:
\[
  \text{RR}_F \;=\; \frac{\Pr \bigl(Y=\hat{Y}\mid F=1\bigr)\;}{\Pr \bigl(Y=\hat{Y}\mid F=0\bigr)\;}
\]
If $\text{RR}_F=1$, it means that there is no change in accuracy. If $\text{RR}_F>1$, it means the accuracy is higher when the feature aligns with the ground truth ($F=1$), and conversely for $\text{RR}_F<1$.
Since RRs are within-model quantities and depend on the model's baseline accuracy when a cue is absent, bar heights should not be compared across models (see \Cref{app:risk-ratio} for details).
For the order cue, we report $\max(\text{RR}_O, 1/\text{RR}_O)$, which captures the effect size irrespective of direction.

\begin{figure}[t]
  \includegraphics[width=\columnwidth]{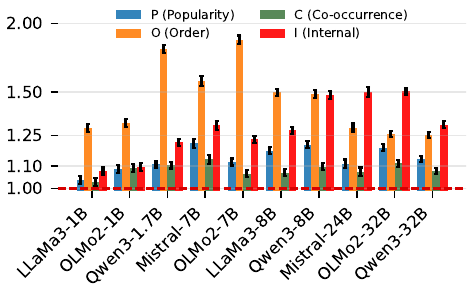}
  \caption{
    Risk ratios (log scale) for each feature, with 95\% confidence intervals from bootstrapping over prompt templates and datasets. See \cref{fig:bos_detailed} in \cref{app:sec:bos_detailed} for per-dataset, per-model results.
  }
  \label{fig:biases}
\end{figure}

\paragraph{Results.} Figure \ref{fig:biases} shows the risk ratios for the different models. Order ($O$) is the dominant shortcut for all models. For most smaller and mid-sized models, $O$ also displays a higher risk ratio than $I$, meaning that the order in which entities are presented has a stronger impact on model predictions than the model's knowledge of the numerical attributes. For example, Qwen3-1.7B reaches an $\text{RR}_O$ of almost $2$, meaning the model is roughly twice as likely to answer correctly
when the larger entity appears in its preferred position.
Popularity ($P$) effects are smaller ($\text{RR} \approx 1.1-1.25$) but are consistently present across model sizes, reflecting a persistent ``fame implies bigger'' heuristic. Co-occurrence ($C$) shows the weakest effect, typically near $\text{RR} \approx 1.05-1.15$, but also remains consistently present across all model sizes. Internal-ground-truth alignment ($I$) grows in importance with scale, reaching RRs around $1.5$ for the largest models. This indicates that when their own extracted numbers agree with reality, they are more likely to answer correctly w.r.t.\ the ground truth compared to smaller models. In summary, we find that while all models are susceptible to biases, larger models tend to rely more on their internal knowledge, whereas smaller models are more influenced by heuristic cues.

\section{Can LLM Predictions Be Explained?}
\label{sec:meta-predictor}

The previous analysis isolated the effect of each cue in turn. Yet in practice, multiple cues may counteract or reinforce each other, raising the question: \emph{can we build a simple model that predicts the LLM's choice better than its own numbers, purely from such surface features?} This section formalizes this idea via a simple \emph{meta-predictor}, a logistic regression model trained to predict whether the LLM will select the first or second entity in a pairwise comparison. The meta-predictor is provided with two binary features, namely whether the first entity is more popular than the other, and whether the first entity is more associated with magnitude descriptors (via cosine similarity in ConceptNet embeddings, see \Cref{sec:biases}). The meta-predictor is trained to predict whether the LLM will choose the first or the second entity, and can thus also take position bias into account. The meta-predictor is trained separately for each model, prompt template, and numerical attribute using $5$-fold cross-validation.

In \Cref{fig:meta_pred}, we contrast the performance of the meta-predictor with a strategy where a model's pairwise predictions always follow their own numerical predictions (absolute accuracies can be found in \Cref{sec:detailed_meta_pred}). For the smallest models, \emph{surface cues predict the model's pairwise choice better than the extracted numerical values}. For larger models, \emph{the extracted numerical values become more predictive (although the meta-predictor remains competitive)}.

\subsection{Fine-grained Analysis}
\label{sec:cases}

\begin{figure}[t]
    \includegraphics[width=\columnwidth]{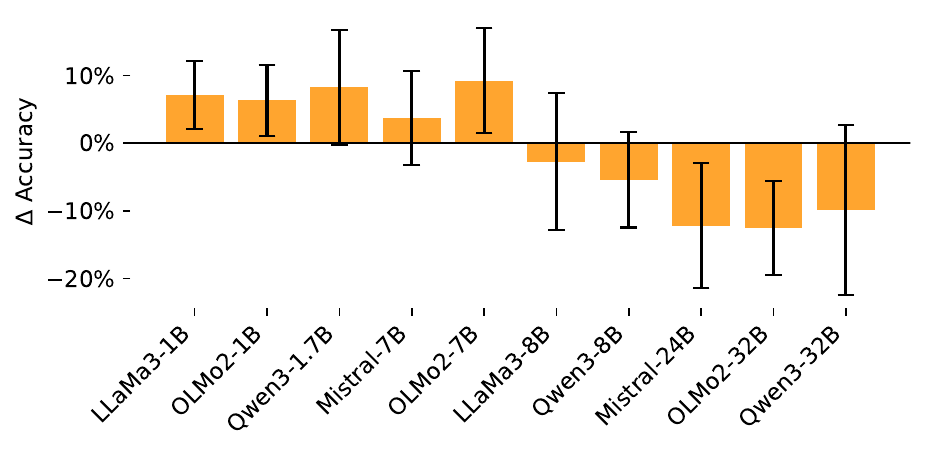}
    \caption{
        For each language model, we report the mean improvement and standard deviation of a logistic-regression meta-predictor that relies solely on three surface cues,  
        relative to a strategy that follows the model's extracted numerical values.
        Positive values indicate that the bias-based predictor anticipates the model's pairwise choice more accurately than the model's own numbers.
    }
    \label{fig:meta_pred}
\end{figure}

To better understand these dynamics,  we classify each test sample into the following cases:
\begin{description}[noitemsep,topsep=0pt,parsep=0pt,partopsep=0pt]
    \item[\colorbox{Case1Color}{\textbf{Case 1}}] Pairwise and numerical predictions agree, meta-predictor disagrees $\Rightarrow$ numerical reasoning
    \item[\colorbox{Case2Color}{\textbf{Case 2}}] All three predictions agree $\Rightarrow$ numerical reasoning or superficial cues
    \item[\colorbox{Case3Color}{\textbf{Case 3}}] Pairwise and meta-predictor agree, numerical prediction disagrees $\Rightarrow$ superficial cues
    \item[\colorbox{Case4Color}{\textbf{Case 4}}] Pairwise prediction disagrees with both numerical and meta-predictor $\Rightarrow$ unexplained / noise
\end{description}

\begin{figure}[t]
    \includegraphics[width=\columnwidth]{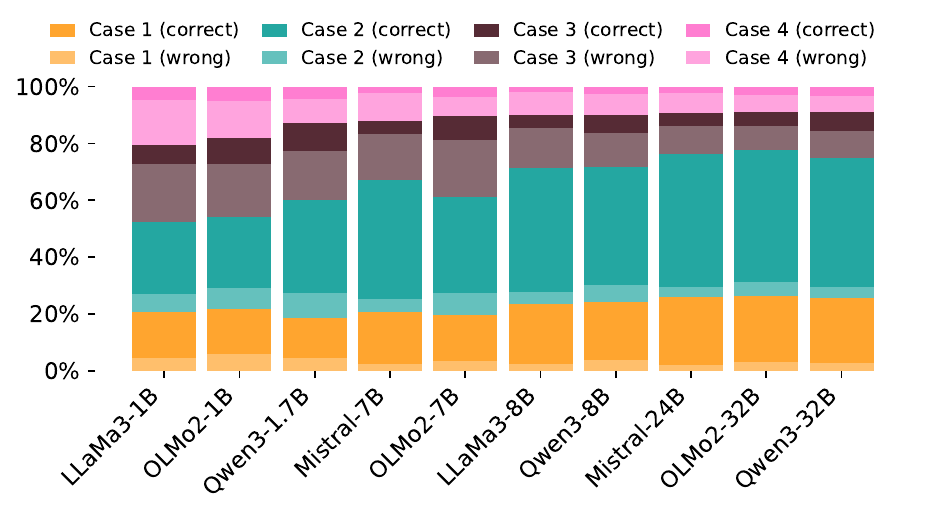}
    \caption{Proportion of the four cases for different models, aggregated over all datasets. For each of the four cases, the upper (darker) segment marks correct predictions, and the lower (lighter) segment marks errors with respect to the ground truth.}
    \label{fig:case_distribution}
\end{figure}

\noindent Figure~\ref{fig:case_distribution} shows the distribution of these cases for the different models, where each case is further split into two sub-cases, depending on whether the prediction matched the ground truth or not.\footnote{A breakdown per dataset can be found in \Cref{app:sec:cases_per_dataset}.}

We can make several key observations.
\begin{inparaenum}
    \item First, most pairwise predictions can be explained by numerical reasoning, surface biases, or both (Cases~\mbox{1--3}), especially in larger models. Case~4, where neither the numerical prediction nor the meta-predictor aligns with the pairwise output, is rare, suggesting that models rely only minimally on unmodeled heuristics or random behavior.
    \item Consistent with our earlier findings from Figure \ref{fig:meta_pred}, we observe a clear difference in how smaller and larger models make decisions. Smaller models such as LLaMa3-1B and OLMo2-1B are more frequently guided by surface-level biases (Case~3) than by their own numerical predictions (Case~1), whereas larger models show the opposite trend, relying more consistently on numerical information.
    \item The case breakdown clarifies how different types of prediction behavior relate to correctness. In Case~2, where all three predictors agree, models are almost always correct.
    \item Case~1 is also associated with high accuracy, whereas Case~3 shows the opposite pattern. This supports the view that the meta-predictor captures surface cues that can lead the model away from correct decisions when they conflict with its numerical knowledge.
    \item Finally, this analysis helps explain a counterintuitive pattern noted earlier: some mid-sized models, including Mistral-7B and LLaMA3-8B, achieve relatively high numerical accuracy but underperform in pairwise comparisons (cf.\ Figure~\ref{fig:general_acc}). The case distribution reveals that these models frequently follow surface heuristics that go against their numerical knowledge (Case~3).
\end{inparaenum}

\subsection{When Do Models Rely on Surface Cues?}
\label{sec:case1_vs_case3}

All models sometimes show signs of being a world model by following their own extracted numbers (Case~1), and sometimes appear to be statistical parrots, following surface cues instead (Case~3). We want to understand \emph{what differentiates these two modes of behavior}. For this analysis, we exclude test examples that fall in Cases 2 and 4, as they might confound interpretation.\footnote{A detailed analysis of Case~2 can be found in \Cref{app:sec:case2_swap}.} For each entity, we consider five metrics that we hypothesize might influence whether a model relies on numerical reasoning (Case~1) or superficial cues (Case~3):
\begin{inparaenum}
    \item \textbf{Ground-truth value} (GT) for the considered numerical attribute;
    \item \textbf{Model-extracted value} (NumEx), i.e., the model's prediction for the considered numerical attribute;
    \item \textbf{Symmetric Mean Absolute Percentage Error} of the extracted numbers relative to ground truth (SMAPE, see \Cref{sec:numex_error});
    \item \textbf{Coefficient of Variation} of the extracted numbers across prompt templates (CV, see \Cref{subsec:numex-cv}), which offers a proxy for the model's confidence in the numerical value;
    \item \textbf{Popularity} (QRank) of the entity.
\end{inparaenum}
Each of the metrics is used to compute two statistics for entity pairs: the \textbf{mean} log-value of the metric across both entities and the \textbf{difference} between their log-values. We use the logarithm of the numerical values, rather than the values themselves, to focus on their order of magnitude. We include both the mean and the difference as they capture different effects: The mean allows us to test, for instance, whether larger values are more common in Case~1 or Case~3 (e.g., whether entity popularity affects how the model makes a prediction). The difference allows us to test, for instance, whether a clearer gap between the model-extracted values is predictive.

For each feature $x$, we want to know whether it tends to be larger in Case 1 or larger in Case 3, and by how much. Cohen's $d$ \cite{cohen_statistical_1988} answers exactly that: it is the difference in group means measured in units of a typical within-group standard deviation (SD), making it unitless and comparable across features.\footnote{Details on how this statistic is computed can be found in \cref{sec:appendix_cohen}.}
In our setting, a positive value $d$ indicates that feature $x$ is, on
average, larger in Case~1, while a negative $d$ indicates it is larger in
Case~3. The magnitude $|d|$ says how strongly the groups differ, measured in
pooled-SD units. For instance, $d = 0.5$ means the average Case 1 value is half
a standard deviation larger than the average Case 3 value. Note that this
analysis is descriptive and does not identify causal effects.

\begin{figure}[t]
    \centering
    \includegraphics[width=\columnwidth]{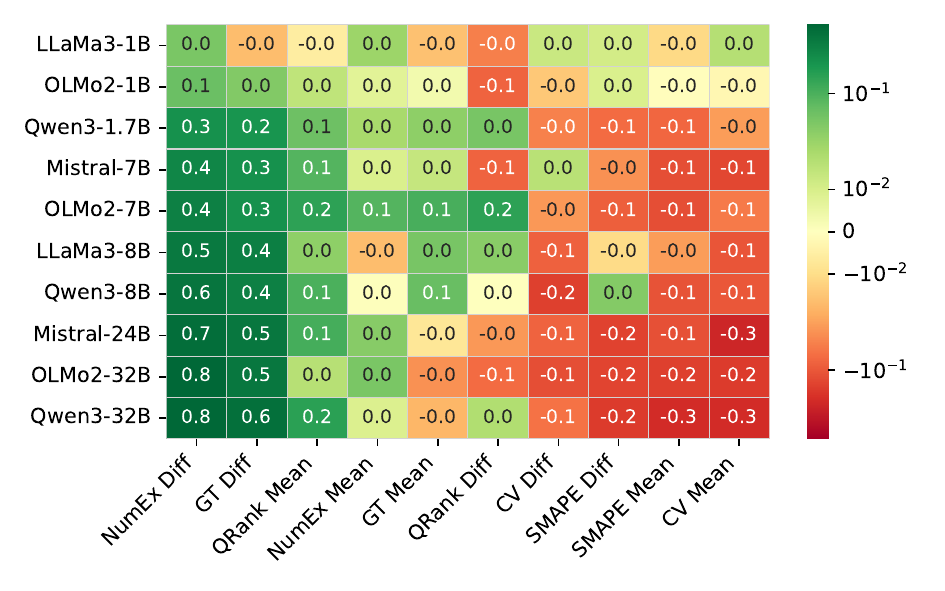}
    \caption{Signed Cohen's $d$ for contrasting Case 1 vs.\ Case 3. Green means the feature value tends to be larger in Case 1 (numerical knowledge), red means that the value tends to be larger in Case 3 (surface cues). Cell labels show the mean effect. }
    \label{fig:case1_vs_case3}
\end{figure}

\Cref{fig:case1_vs_case3} shows the $d$ statistic for each of the 10 features, per model, aggregated over all the datasets.\footnote{A detailed breakdown can be found in \Cref{app:sec:case1_vs_3}.} Several clear regularities can be observed. First, the NumEx difference and the GT difference are both larger for Case 1 than for Case 3. Large differences intuitively mean that relying on the numerical attributes is safer, as even highly approximate numerical knowledge is sufficient for making reliable pairwise predictions in such cases. Interestingly, this effect is much more pronounced for the larger models. For instance the $d$-statistic for NumEx-diff rises from $\approx{0.04}$ at 1B to $\approx{0.79}$ at 32B. Conversely, SMAPE and CV means tend to be much higher for Case 3 than for Case 1. High values for these features indicate that the model's knowledge of the required numerical attributes is noisy, which makes relying on them riskier. Using alternative heuristics may thus be a rational choice in such cases. Again, we see that this effect is most pronounced for the largest models. Overall, our analysis thus supports the view that larger models make more principled choices when deciding between the two strategies (i.e., relying on numerical attributes vs. heuristic cues).

\section{How does CoT Affect Predictions?}
\label{sec:thinking}

\begin{figure}[t]
        \centering
        \includegraphics[width=\columnwidth]{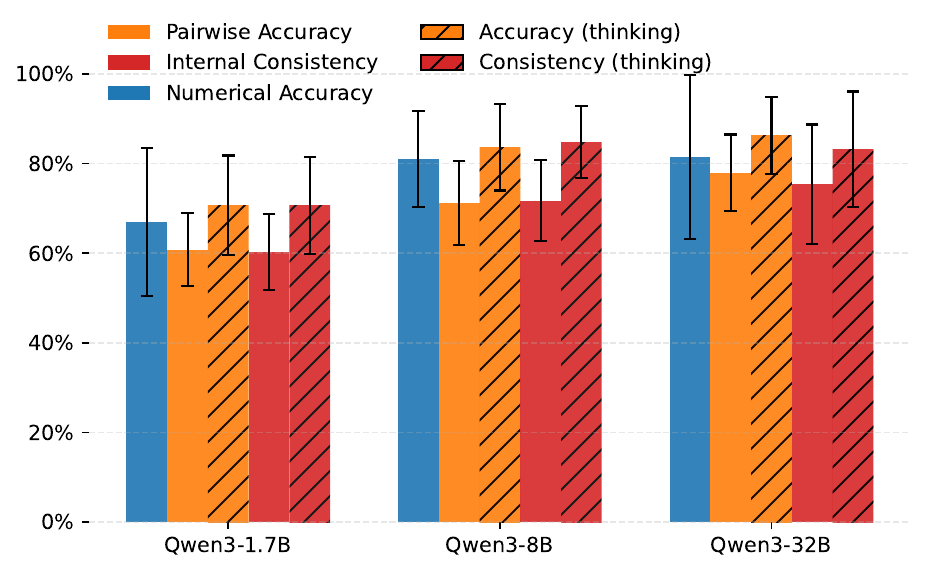}
        \caption{Comparison of the performance with and without thinking, in terms of pairwise accuracy and internal consistency. The figure also reports numerical accuracy (without thinking). Error bars show the standard deviation over datasets.}
        \label{fig:thinking_acc}
\end{figure}

In this section, we investigate whether prompting LLMs to ``think'' before answering improves the internal consistency of their pairwise numerical comparisons. We ask: \emph{does explicit reasoning help models use their own numerical predictions more faithfully when making pairwise judgments?} To address this question, we focus on the models from the Qwen3 series, which have been fine-tuned specifically for reasoning tasks using chain-of-thought supervision. During inference, we permit the model to generate up to $1024$ new tokens, ensuring that the ``thinking'' prompts have a large enough budget to verbalize intermediate steps. We use the same prompt templates as before, but append a start-of-thought token at the end.

Figure \ref{fig:thinking_acc} shows that prompting the Qwen3 models to ``think'' before responding yields a clear performance lift. On average, pairwise accuracy rises by 5--9 percentage points, and internal consistency rises \emph{by a nearly identical margin}. Intuitively, we might expect a chain-of-thought prompt to first retrieve both numbers and then compare them mechanically, pushing internal consistency to nearly 100\%. We see that this is not the case. To better understand these results, we manually inspected a sample of chain-of-thought traces produced by the Qwen3 models. The analysis reveals several recurring tendencies. In many cases, the model appears to make up its mind before retrieving any numbers, then generates numerical statements that merely serve to justify the chosen answer, reinforcing the view that chain-of-thought traces should not be treated as faithful evidence about whether the model relied on numerical reasoning or heuristics \cite{DBLP:conf/nips/0002G0024, lyu-etal-2023-faithful, lanham_measuring_2023, paul-etal-2024-making, chen_reasoning_2025}. In other traces, the retrieved numbers differ from those obtained when the model is asked for the values directly, sometimes being closer to the ground truth but frequently inaccurate or inconsistent. We did notice more generally that different prompts sometimes lead to different numerical values.\footnote{An in-depth analysis of this can be found in \Cref{subsec:numex-cv}.} In other cases, the reasoning step is skipped altogether, with the model producing a direct answer despite the thinking prompt. When reasoning occurs, it sometimes relies on heuristic arguments. Finally, for some samples, the model used more than $1024$ tokens in the thought process, therefore not yielding an answer. 

Together, these observations explain why chain-of-thought prompting improves pairwise accuracy and internal consistency without eliminating inconsistency.
The reasoning traces often reflect rationalization rather than deliberate computation, and the modest gains likely stem from occasional improvements in number retrieval or from semantically plausible heuristics that happen to yield the correct answer.
A more systematic analysis of these reasoning patterns, and their relation to numerical faithfulness, is left for future work (see \cref{sec:app:detailed_cot} for representative examples).

\section{Related Work}
\label{sec:related_work}

Previous work has already found that LLM predictions can be influenced by various types of superficial features.
\citet{wang-etal-2024-large-language-models-fair} identified a position bias in LLM evaluators, where the result is influenced by the order in which candidates are presented. 
\citet{DBLP:journals/corr/abs-2309-13638} found how the accuracy of an LLM is influenced by the probability of the output, which aligns with our findings of popularity bias.
The fact that shallow co-occurrence statistics can mislead LLMs, being the third bias that we study, has also been shown in several studies \cite{kang-choi-2023-impact}.
While it is thus not surprising that these biases are present in our analysis, the significance of our finding stems from the extent to which these biases affect the result. 
The lack of internal consistency of LLMs with numerical features also aligns with various findings from the literature.
In the context of ranking, the non-transitive nature of pairwise judgments by LLMs has been highlighted \cite{DBLP:journals/corr/abs-2502-14074,kumar-etal-2024-ranking}. 
The reversal curse \cite{DBLP:conf/iclr/BerglundTKBSKE24}, where models fail to answer inverse formulations of questions, also suggests a lack of internal consistency.
Along similar lines, \citet{he2025supposedly} report systematic asymmetries in recognizing logically equivalent relational facts, which they trace back to entity-frequency imbalances in the pre-training data (high-frequency subjects paired with low-frequency objects are recognized more reliably than the inverse). 
\citet{DBLP:conf/icml/Allen-ZhuL24} find that LLMs sometimes memorize knowledge without being capable of reliably exploiting it for answering questions.
The compositionality gap, where models can answer individual sub-questions but fail to compose them into correct multi-hop answers, has been documented by \citet{press-etal-2023-measuring}, who found that scaling improves single-hop performance faster than multi-hop performance. This parallels our finding that models possess numerical knowledge but fail to reliably apply it in pairwise comparisons.
The problem of ranking entities with LLMs was studied by \citet{kumar-etal-2024-ranking}, but their focus was on designing fine-tuning strategies.
Regarding calibration, i.e., how well models know what they don't know, \citet{kadavath_language_2022} demonstrated that larger models are more aware of their own knowledge boundaries, with scale playing a crucial role. Our work extends these insights by showing that larger models not only know what they know, but can also strategically choose when to rely on that knowledge versus when to use heuristics.

\section{Conclusion}
\label{sec:conclusion}
We have analyzed how LLMs behave when asked to compare entities along some numerical attribute.
Intuitively, an LLM could simply extract the attribute values for the two given entities and compare these.
However, we found their actual performance dramatically underperforms such a strategy. 
Our experiments suggest that LLMs switch between two strategies: a principled approach based on their knowledge of the numerical attributes and a heuristic approach based on surface cues, such as entity popularity, co-occurrence statistics, and the ordering of the entities in the prompt. Furthermore, we found that larger models tend to choose between these strategies in a more principled way, being more likely to rely on numerical attributes when their numerical knowledge is more (likely to be) reliable. Finally, in our experiments with CoT-based reasoning models, we found that predictions align better, but still not perfectly, with the models' numerical knowledge.

Our findings offer a nuanced perspective on the ongoing debate between the world-model and statistical-parrot views of LLMs. Rather than supporting either extreme, our results suggest that LLMs operate in a hybrid manner: they possess genuine world knowledge (numerical attributes) but do not always deploy it consistently.
Importantly, the ability to strategically select between knowledge-driven reasoning and heuristic shortcuts emerges with scale, suggesting that larger models are developing a form of meta-cognitive capability. Our work thus provides a first step towards a more sophisticated understanding of LLM behavior: one where the question is not whether models understand or merely parrot, but rather \emph{when} and \emph{how} they choose between different reasoning strategies.

\section*{Acknowledgments}
Jae Hee Lee was supported by the Deutsche Forschungsgemeinschaft (DFG, German Research Foundation) – project number 551629603.
Steven Schockaert was supported by EPSRC grant EP/W003309/1.

\section*{Limitations}
\label{sec:limitations}

Our study has been limited to an analysis of the outputs of LLMs, and we have not attempted to interpret these models mechanistically. For instance, it would be interesting to see whether (or under which conditions) updating the numerical knowledge inside models would alter their pairwise judgments.
Furthermore, our analysis has been limited to zero-shot (chain-of-thought) prompting. In preliminary experiments, we observed that few-shot prompting may help to partially overcome some of the biases that we studied, although not entirely. Similarly, it would be interesting to study whether the biases persist after fine-tuning models on ranking tasks.
Lastly, we study pairwise comparisons for numerical attributes in a controlled setting, which yields objective ground truth and orthogonal cue manipulation, but our findings may not fully transfer to open-domain question answering, non-numerical attributes, multi-entity ranking, or multi-hop reasoning.

\bibliography{references, ref2}

\clearpage
\appendix
\crefalias{section}{appendix}
\crefalias{subsection}{appendix}
\crefalias{subsubsection}{appendix}
\section{Prompts}
\label{sec:prompts}

\Cref{tab:pairwise_prompts} and \Cref{tab:numerical_prompts} list the prompt templates used in our experiments. Each attribute-dataset combination includes six pairwise prompts (three prompting for the ``larger'' entity and three for the ``smaller'' one) and three numerical extraction prompts.
As a system prompt, we used the following:
\begin{quote}
    \textit{You are a chatbot to help with general knowledge questions. You answer as short and concise as possible. Meaning, you should not provide more information than what is asked for. If you are asked to compare two entities answer with the name of the correct one only.}
\end{quote} 

\begin{table*}[p]
    \centering
    \scriptsize    
    \begin{tabularx}{\textwidth}{@{}lX@{}}
        \toprule
        Entity Type                      & Prompt                                                                                                                                                       \\
        \midrule
        \multirow{6}{*}{Atoms}           & Answer with the one name only. Which chemical element has the higher atomic number? \{entity1\} or \{entity2\}?                                              \\
                                         & Please state the chemical element with the higher atomic number only. \{entity1\} or \{entity2\}?                                                            \\
                                         & Answer only with the correct name. Which chemical element has a higher number of protons? \{entity1\} or \{entity2\}?                                        \\
                                         & Answer with the one name only. Which chemical element has the lower atomic number? \{entity1\} or \{entity2\}?                                               \\
                                         & Please state the chemical element with the lower atomic number only. \{entity1\} or \{entity2\}?                                                             \\
                                         & Answer only with the correct name. Which chemical element has a lower number of protons? \{entity1\} or \{entity2\}?                                         \\
        \midrule
        \multirow{6}{*}{Buildings}       & Only state the name of the taller building. Which building is taller? \{entity1\} or \{entity2\}?                                                            \\
                                         & Respond with only the name of the taller building. Which building is taller? \{entity1\} or \{entity2\}?                                                     \\
                                         & Provide only the name of the taller building. Which building is taller? \{entity1\} or \{entity2\}?                                                          \\
                                         & Only state the name of the shorter building. Which building is shorter? \{entity1\} or \{entity2\}?                                                          \\
                                         & Respond with only the name of the shorter building. Which building is shorter? \{entity1\} or \{entity2\}?                                                   \\
                                         & Provide only the name of the shorter building. Which building is shorter? \{entity1\} or \{entity2\}?                                                        \\
        \midrule
        \multirow{6}{*}{Cities}          & Only state the name of the more populous city. Which city has a larger population? \{entity1\} or \{entity2\}?                                               \\
                                         & Respond with only the name of the more populous city. Which city has a larger population? \{entity1\} or \{entity2\}?                                        \\
                                         & Provide only the name of the more populous city. Which city has a larger population? \{entity1\} or \{entity2\}?                                             \\
                                         & Only state the name of the less populous city. Which city has a smaller population? \{entity1\} or \{entity2\}?                                              \\
                                         & Respond with only the name of the less populous city. Which city has a smaller population? \{entity1\} or \{entity2\}?                                       \\
                                         & Provide only the name of the less populous city. Which city has a smaller population? \{entity1\} or \{entity2\}?                                            \\
        \midrule
        \multirow{6}{*}{Countries}       & Only state the name of the more populous country. Which country has a larger population? \{entity1\} or \{entity2\}?                                         \\
                                         & Respond with only the name of the more populous country. Which country has a larger population? \{entity1\} or \{entity2\}?                                  \\
                                         & Provide only the name of the more populous country. Which country is more populous? \{entity1\} or \{entity2\}?                                              \\
                                         & Only state the name of the less populous country. Which country has a smaller population? \{entity1\} or \{entity2\}?                                        \\
                                         & Respond with only the name of the less populous country. Which country has a smaller population? \{entity1\} or \{entity2\}?                                 \\
                                         & Provide only the name of the less populous country. Which country is less populous? \{entity1\} or \{entity2\}?                                              \\
        \midrule
        \multirow{6}{*}{Mountains}       & Only state the name of the higher mountain. Which mountain is higher? \{entity1\} or \{entity2\}?                                                            \\
                                         & Respond with only the name of the mountain that has a greater elevation. Which mountain stands taller? \{entity1\} or \{entity2\}?                           \\
                                         & Provide only the name of the higher mountain. Which mountain has a greater elevation? \{entity1\} or \{entity2\}?                                            \\
                                         & Only state the name of the lower mountain. Which mountain is lower? \{entity1\} or \{entity2\}?                                                              \\
                                         & Respond with only the name of the mountain that has a lesser elevation. Which mountain stands lower? \{entity1\} or \{entity2\}?                             \\
                                         & Provide only the name of the lower mountain. Which mountain has a smaller elevation? \{entity1\} or \{entity2\}?                                             \\
        \midrule
        \multirow{6}{*}{Peppers}         & Only state the name of the hotter pepper. Which pepper has a higher Scoville Heat Unit rating? \{entity1\} or \{entity2\}?                                   \\
                                         & Respond with only the name of the hotter pepper. Which pepper is spicier based on Scoville Heat Units? \{entity1\} or \{entity2\}?                           \\
                                         & Provide only the name of the hotter pepper. Which pepper has the greater spiciness level according to the Scoville scale? \{entity1\} or \{entity2\}?        \\
                                         & Only state the name of the milder pepper. Which pepper has a lower Scoville Heat Unit rating? \{entity1\} or \{entity2\}?                                    \\
                                         & Respond with only the name of the milder pepper. Which pepper is less spicy based on Scoville Heat Units? \{entity1\} or \{entity2\}?                        \\
                                         & Provide only the name of the milder pepper. Which pepper has a lower spiciness level according to the Scoville scale? \{entity1\} or \{entity2\}?            \\
        \midrule
        \multirow{6}{*}{People (social)} & Only state the name of the person with more social media followers. Which person has a larger social media following? \{entity1\} or \{entity2\}?            \\
                                         & Respond with only the name of the individual who has more social media followers. Between \{entity1\} and \{entity2\}, who has a larger following?           \\
                                         & Provide only the name of the person with more social media followers. Who has a larger social media following? \{entity1\} or \{entity2\}?                   \\
                                         & Only state the name of the person with fewer social media followers. Which person has a smaller social media following? \{entity1\} or \{entity2\}?          \\
                                         & Respond with only the name of the individual who has fewer social media followers. Between \{entity1\} and \{entity2\}, who has a smaller following?         \\
                                         & Provide only the name of the person with fewer social media followers. Who has a smaller social media following? \{entity1\} or \{entity2\}?                 \\
        \midrule
        \multirow{6}{*}{Rivers}          & Only state the name of the longer river. Which river is longer? \{entity1\} or \{entity2\}?                                                                  \\
                                         & Respond with only the name of the longer river. Which river extends further? \{entity1\} or \{entity2\}?                                                     \\
                                         & Provide only the name of the river with the longer course. Which of these rivers covers a longer distance? \{entity1\} or \{entity2\}?                       \\
                                         & Only state the name of the shorter river. Which river is shorter? \{entity1\} or \{entity2\}?                                                                \\
                                         & Respond with only the name of the shorter river. Which river extends a shorter distance? \{entity1\} or \{entity2\}?                                         \\
                                         & Provide only the name of the river with the shorter course. Which of these rivers covers a shorter distance? \{entity1\} or \{entity2\}?                     \\
        \midrule
        \multirow{6}{*}{Stadiums}        & Only state the name of the stadium with a larger seating capacity. Which stadium can accommodate more spectators? \{entity1\} or \{entity2\}?                \\
                                         & Respond with only the name of the stadium that has a greater seating capacity. Which stadium has more seats? \{entity1\} or \{entity2\}?                     \\
                                         & Provide only the name of the stadium with a higher capacity. Which stadium can hold more people? \{entity1\} or \{entity2\}?                                 \\
                                         & Only state the name of the stadium with a smaller seating capacity. Which stadium can accommodate fewer spectators? \{entity1\} or \{entity2\}?              \\
                                         & Respond with only the name of the stadium that has a lower seating capacity. Which stadium has fewer seats? \{entity1\} or \{entity2\}?                      \\
                                         & Provide only the name of the stadium with a smaller capacity. Which stadium can hold fewer people? \{entity1\} or \{entity2\}?                               \\
        \midrule
        \multirow{6}{*}{Universities}    & Only state the name of the university with more enrolled students. Which university has a larger student population? \{entity1\} or \{entity2\}?             \\
                                         & Respond with only the name of the university that has a greater number of students. Which university has more students enrolled? \{entity1\} or \{entity2\}? \\
                                         & Provide only the name of the university with a higher student enrollment. Which university has the largest student body? \{entity1\} or \{entity2\}?         \\
                                         & Only state the name of the university with fewer enrolled students. Which university has a smaller student population? \{entity1\} or \{entity2\}?           \\
                                         & Respond with only the name of the university that has a lower number of students. Which university has fewer students enrolled? \{entity1\} or \{entity2\}?  \\
                                         & Provide only the name of the university with a lower student enrollment. Which university has the smallest student body? \{entity1\} or \{entity2\}?         \\
        \bottomrule
    \end{tabularx}
\caption{Pairwise prompts for all entity types.}
    \label{tab:pairwise_prompts}
\end{table*}

\begin{table*}[p]
    \centering
    \scriptsize    
    \begin{tabularx}{\textwidth}{@{}lX@{}}
        \toprule
        Entity Type                      & Prompt                                                                                                                                   \\
        \midrule
        \multirow{3}{*}{Atoms}           & What is the atomic number of \{entity\}?                                                                                                 \\
                                         & Please state the atomic number of \{entity\}.                                                                                            \\
                                         & How many protons does \{entity\} have?                                                                                                   \\
        \midrule
        \multirow{3}{*}{Buildings}       & What is the height of the building \{entity\} in meters?                                                                                 \\
                                         & How tall is the \{entity\} building in meters?                                                                                           \\
                                         & Please state the height of the \{entity\} building measured in meters?                                                                   \\
        \midrule
        \multirow{3}{*}{Cities}          & What is the population size of \{entity\}, including its metropolitan area?                                                              \\
                                         & What is the total population of \{entity\}, encompassing its metropolitan region?                                                        \\
                                         & Please state the population of \{entity\}, including its metropolitan area.                                                              \\
        \midrule
        \multirow{3}{*}{Countries}       & What is the population size of the country \{entity\} in 2023?                                                                           \\
                                         & What is the number of inhabitants in \{entity\} as of 2023?                                                                              \\
                                         & Please state the population of \{entity\} in 2023.                                                                                       \\
        \midrule
        \multirow{3}{*}{Mountains}       & What is the height of \{entity\} in meters above sea level?                                                                              \\
                                         & What is the altitude of \{entity\} expressed in meters above sea level?                                                                  \\
                                         & Please state the height of \{entity\} in meters above sea level.                                                                         \\
        \midrule
        \multirow{3}{*}{Peppers}         & What is the Scoville Heat Unit (SHU) rating of the \{entity\} pepper?                                                                    \\
                                         & How spicy is the \{entity\} pepper in terms of Scoville Heat Units?                                                                      \\
                                         & Please state the Scoville Heat Unit value of the \{entity\} pepper.                                                                      \\
        \midrule
        \multirow{3}{*}{People (social)} & Do not list multiple platforms! Only answer with a single number. How many social media followers does \{entity\} have across platforms? \\
                                         & Provide only the total number of social media followers for \{entity\} across all platforms.                                             \\
                                         & How many social media followers does \{entity\} have in total? Answer with a single number across all platforms.                         \\
        \midrule
        \multirow{3}{*}{Rivers}          & What is the length of the \{entity\} river in km?                                                                                        \\
                                         & How many kilometers long is the \{entity\} river?                                                                                        \\
                                         & Can you provide the length of the \{entity\} river in kilometers?                                                                        \\
        \midrule
        \multirow{3}{*}{Stadiums}        & What is the seating capacity of the \{entity\} stadium?                                                                                  \\
                                         & How many spectators can the \{entity\} stadium accommodate?                                                                              \\
                                         & Please state the total number of seats available in the \{entity\} stadium.                                                              \\
        \midrule
        \multirow{3}{*}{Universities}    & How many students are enrolled at \{entity\}?                                                                                            \\
                                         & What is the total student enrollment at \{entity\}?                                                                                      \\
                                         & Please state the number of students enrolled at \{entity\}.                                                                              \\
        \bottomrule
    \end{tabularx}
    \caption{Numerical prompts for all entity types.}
    \label{tab:numerical_prompts}
\end{table*}
\section{Response Parsings}
\label{sec:parsing_details}

To convert free-form model outputs into structured predictions, we use two deterministic regex-based parsing pipelines, one for numerical-value prompts and one for pairwise-comparison prompts. These are explained in detail in the following paragraphs.

\subsection{Numerical Prompts}
% For the pointwise paradigm, 
The following parsing procedure is applied. For attributes where a physical unit is expected (e.g., meters, kilometers), we begin by extracting all numeric values from the response and, if applicable, convert them into the unit requested by the prompt. For instance, if the model returns a distance in kilometers when meters were asked for, we apply the appropriate conversion factor. If the unit is missing or ambiguous, we assume the number is in the expected unit. We also normalize magnitude modifiers such as ``\textit{k}'', ``\textit{million}'', etc. (e.g., $60\text{k} \rightarrow 60,000$). If multiple valid numbers are found, we select the one closest to the ground truth value, under the assumption that the model may have approximated the correct answer. We do this because the models sometimes responded with the year when its knowledge cutoff was, for example: ``\textit{In 2023, XY had a population of\ldots}'' or named multiple units. If exactly one number is found, we return it. If no number can be extracted, the response is marked as \texttt{unknown}.

\subsection{Pairwise Prompts}
For the pairwise paradigm, we need to determine which of the two entities was chosen by the model. To do this, we follow a multistep procedure. First, we check whether exactly one of the entity names appears verbatim in the output. If so, we treat that entity as the model's prediction. If both names appear, we search for indicative phrasings that suggest a directional comparison, i.e., statements that clearly identify one entity as having a higher or lower value. If this fails, we check whether the response contains an unambiguous substring of one of the entity names. This accounts for answers like ``China'' when the full name is ``People's Republic of China''. If still unresolved, we apply fuzzy matching\footnote{\url{https://github.com/seatgeek/thefuzz}} to detect typographical or lexical variants. If no reliable match can be made by any of these steps, the response is flagged as \texttt{unknown}.

\subsection{Parsing and Model Failures}
Across all settings, parsing or model failures are rare. \Cref{tab:parsing_fail_rates_summary} reports mean fail rates across datasets for each model, along with the minimum and maximum rates observed across datasets.

\begin{table*}
    \centering
    \footnotesize
\begin{tabular}{llll}
\toprule
 & Numerical Fail Rate & Pairwise Fail Rate & Combined Fail Rate \\
Model &  &  &  \\
\midrule
LLaMa3-1B & 3.7 (0.0–19.2) & 2.1 (0.0–15.8) & 5.8 (0.0–19.3) \\
OLMo2-1B & 0.0 (0.0–0.0) & 0.2 (0.0–0.4) & 0.2 (0.0–0.4) \\
Qwen3-1.7B & 0.4 (0.0–2.8) & 0.0 (0.0–0.1) & 0.4 (0.0–2.8) \\
Mistral-7B & 0.8 (0.0–4.8) & 5.5 (1.0–14.4) & 6.2 (1.7–15.2) \\
OLMo2-7B & 0.0 (0.0–0.3) & 0.3 (0.0–2.2) & 0.4 (0.0–2.2) \\
LLaMa3-8B & 11.4 (0.0–47.6) & 0.0 (0.0–0.1) & 11.4 (0.0–47.6) \\
Qwen3-8B & 0.4 (0.0–2.0) & 0.6 (0.0–5.4) & 0.9 (0.0–5.4) \\
Mistral-24B & 0.1 (0.0–0.3) & 0.1 (0.0–0.3) & 0.1 (0.0–0.5) \\
OLMo2-32B & 2.0 (0.0–18.0) & 0.0 (0.0–0.0) & 2.0 (0.0–18.0) \\
Qwen3-32B & 0.2 (0.0–1.6) & 0.0 (0.0–0.1) & 0.2 (0.0–1.6) \\
Qwen3-1.7B (thinking) & 0.4 (0.0–2.8) & 1.1 (0.2–2.2) & 1.4 (0.2–3.1) \\
Qwen3-8B (thinking) & 0.4 (0.0–2.0) & 1.6 (0.1–4.0) & 1.9 (0.1–4.7) \\
Qwen3-32B (thinking) & 0.2 (0.0–1.6) & 1.4 (0.1–4.0) & 1.6 (0.1–4.0) \\
\bottomrule
\end{tabular}
\caption{Parsing fail rates aggregated across datasets. Each entry reports the mean failure rate (in \%) across datasets for a given model, with the minimum and maximum rates across datasets between parentheses. “Numerical” refers to failures of the numeric-value parsing pipeline, “Pairwise” to failures of the entity-choice parsing pipeline, and “Combined” to failures of either pipeline.}
\label{tab:parsing_fail_rates_summary}
\end{table*}
\section{Prompt Sensitivity}
\label{sec:prompt-sensitivity}

In this section we analyze how sensitive the evaluated models are to the wording of the prompt. For a perfect world model, logically equivalent phrasings would yield identical behavior. But other studies have revealed that LLMs are sensitive to the precise wording of a prompt \cite{brucks_prompt_2025, jiang_how_2020,DBLP:conf/aaai/BouraouiCS20}. Even minor re-phrasings can nudge the model toward a different number, flip a pairwise preference, or introduce noise. To investigate this fragility we answer three questions about prompt sensitivity:

\begin{itemize}
    \item When asking the model for a number, do three paraphrases yield the same number and if not, how different are they?
    \item When asking the model to compare two entities, does the accuracy change when asking for smaller rather than the larger one?
    \item In pairwise comparison, within the same polarity, how often does the model agree on the same answer, when using different prompt variations?
\end{itemize}

\noindent Taken together these three dimensions provide a comprehensive picture of prompt sensitivity from different angles. The following sections unpack each dimension in turn.

%--------------------------------------------------------------------

%--------------------------------------------------------------------

\subsection{Numerical Extraction}
\label{subsec:numex-cv}

\begin{figure*}[p]
    \centering
    \includegraphics[width=\textwidth]{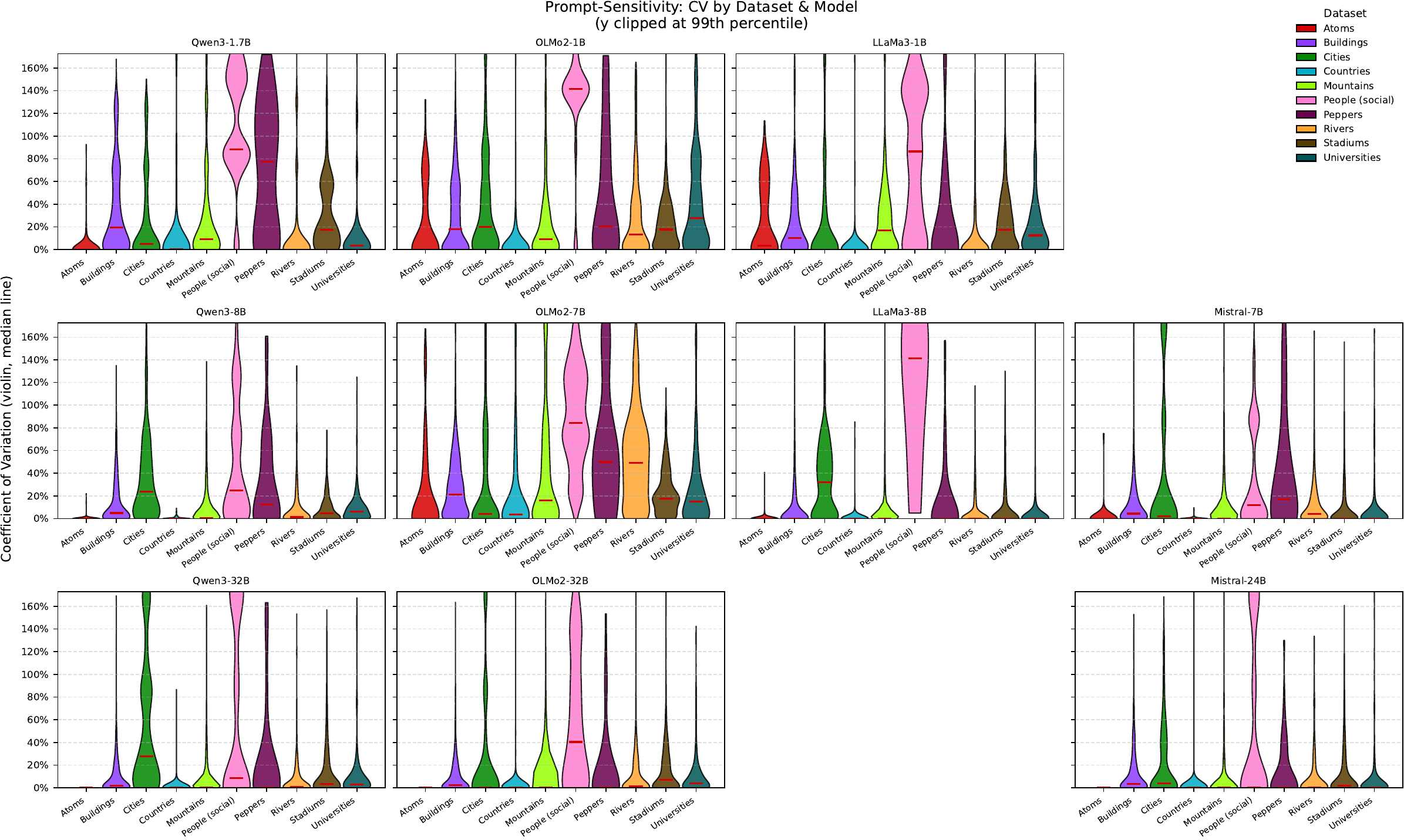}
    \caption{\textbf{Numerical-extraction stability per model and dataset.}
        Violin width encodes the density of row-wise CV values, the horizontal
        bar marks the median.  Rows correspond to model-size buckets
        (\textit{small}, \textit{medium}, \textit{big}); columns to model
        families.}
    \label{fig:app_cv_detail}
\end{figure*}

For each question we ask the model, we use three differently worded numerical-extraction prompts and measure how tightly the answers cluster. We would expect these to be identical for a perfect world-model. Concretely, we take all the extracted values for an entity based on the three different prompts templates and compute the coefficient of variation
\[
    \mathrm{CV}=\sigma/\mu, \qquad \sigma=\text{st.\,dev.},\; \mu=\text{mean}.
\]
CV measures the spread of the three answers around their mean. A low CV means that the three answers are close to one another or even identical (CV=0).
A CV of $0.05$ means that the standard deviation is 5\% of the mean, which is still a tight cluster. A CV of $0.4$, by contrast, indicates a wide spread of answers. The models response differs by about half of the average magnitude of the answers. This would for example be the case if the model outputs $500$, $1000$ and $1500$ for the three different prompts. From a statistical parrot perspective, we would expect a high CV, as the model is likely to pick up on different keywords in the prompt and produce different numbers. To handle cases where the model fails to output any number, we fill the missing values with $0$, thereby penalizing non-numeric answers, when other prompts did yield a number. This is calculated for every entity of every dataset for every model. The results are plotted in \Cref{fig:app_cv_detail} in violin plots with median lines.
Smaller models tend to have higher CV values, as indicated by the median and the shape of the violin. This stands for a greater sensitivity to prompt wording and a higher likelihood of producing varied responses. Generally this trend decreases with model size.

\subsection{Pairwise Ranking}
\label{subsec:pairwise}

We now turn from the stability of extracted numbers to the robustness of direct pairwise decisions under changes in wording, distinguishing between inter- and intra-polarity effects. Intra-polarity refers to the answers of the model when the polarity of the prompt is held constant, e.g., when asking “Which city is larger?” in three different ways. Inter-polarity, by contrast, refers to the comparison of answers when the polarity is flipped, e.g., comparing “Which city is larger?” to “Which city is smaller?”. In principle, a perfect world model would be entirely insensitive to prompt re-phrasings and yield the same pairwise preference regardless of the wording of the prompt template or whether the question asks for the larger or smaller entity. Note that for the experiments in the main paper we consider all prompt templates and entity orderings, i.e., if we had a single pair and therefore twelve different prompts (six templates, two orderings), we would have twelve pairwise decisions to consider. If the model answered 6 of them correctly, while failing to do so in the other 6, we would say that the model has a 50\% accuracy.

\subsubsection{Inter-polarity}
\label{subsubsec:inter-polarity}
\begin{sidewaysfigure*}[p]
    \centering
    \includegraphics[width=\textwidth]{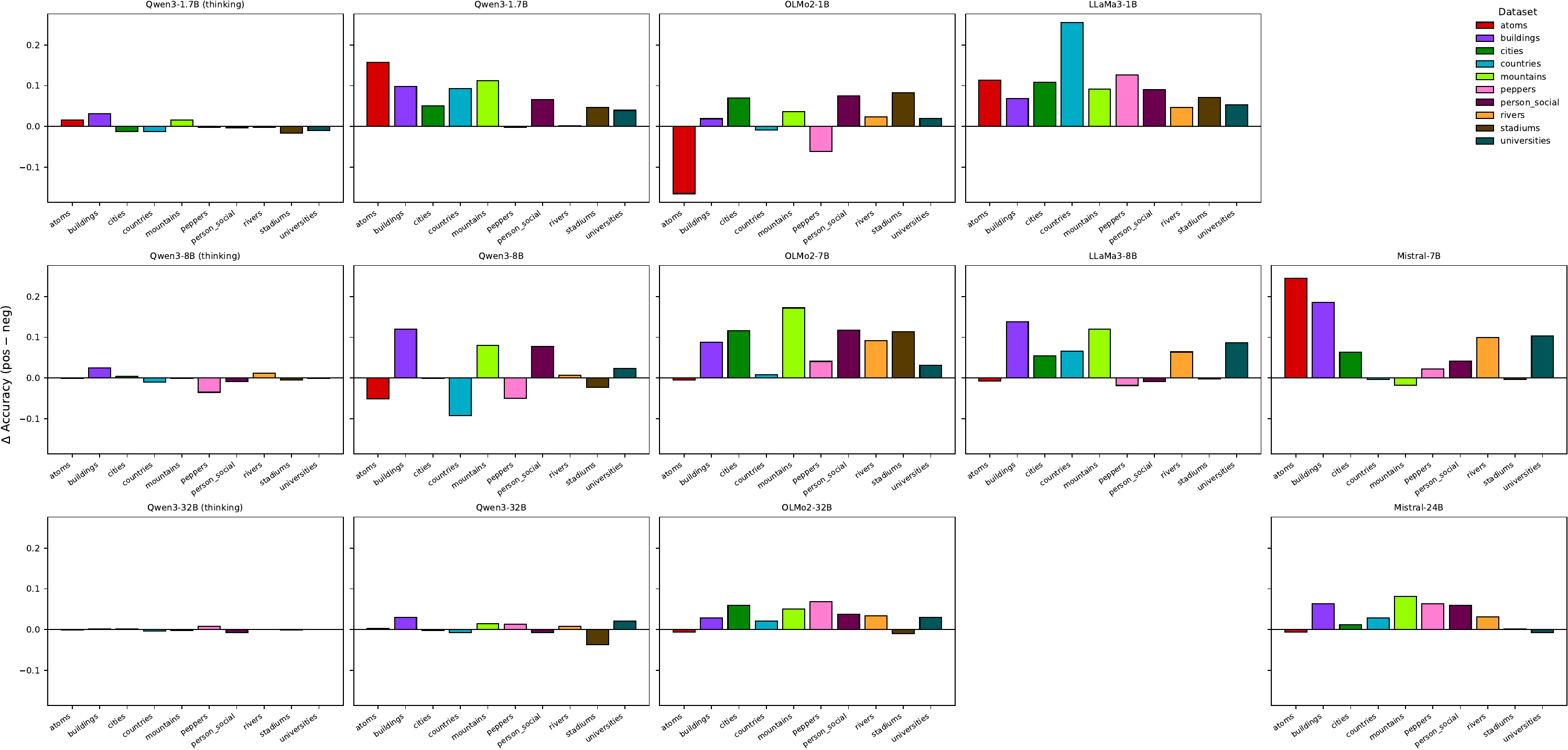}
    \caption{\textbf{Inter-polarity accuracy gap $\Delta\!\text{Acc}$ in detail.}
    Each panel is a single model; bars show the difference
    $\text{Acc}_{\text{Positive Polarity}}-\text{Acc}_{\text{Negative Polarity}}$
    for every dataset (colour legend shared across figures).}
    \label{fig:app_delta_acc_detail}
\end{sidewaysfigure*}

A stable model should be robust to changes in the semantic polarity of the prompt, i.e., if the model can correctly identify the larger entity, it should also be able to identify the smaller one. To test this, we compare the accuracy of the three \emph{larger-than} templates to the accuracy of the three \emph{smaller-than} templates. Flipping the semantic polarity of every template, e.g.\ from \emph{“Which city is \underline{larger}\dots”} to \emph{“Which city is \underline{smaller}\dots”}, yields the accuracy difference \(\Delta\!\text{Acc} = \text{Acc}_{larger} - \text{Acc}_{smaller},\) shown in \Cref{fig:app_delta_acc_detail}.
% Findings
Across the board models prefer the positive polarity: accuracies tend to drop when the question asks for the smaller entity. The gap shrinks with scale. For Qwen3-32B the difference is negligible.

\subsubsection{Intra-polarity}
\label{subsubsec:intra-polarity}

\begin{sidewaysfigure*}[t]
    \centering
    \includegraphics[width=\textwidth]{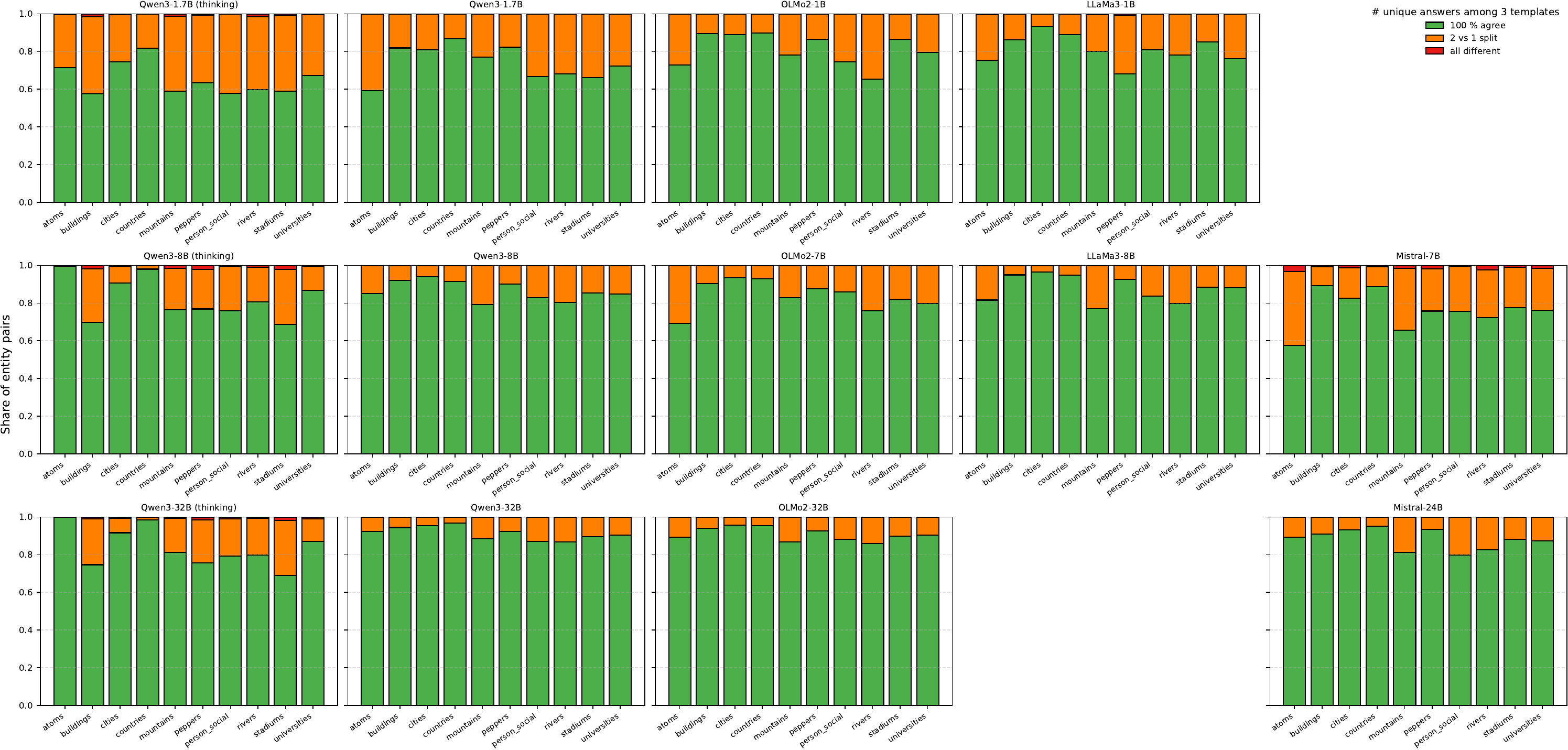}
    \caption{\textbf{Template-majority ratio (TMR) per model and dataset.}
        Stacked bars sum to 100\,\%.
        Green = all three templates agree;
        orange = 2-vs-1 split;
        red = three different winners.}
    \label{fig:app_tmr_detail}
\end{sidewaysfigure*}

Knowing that there is a polarity bias, we now turn to the question of how often the different templates agree with one another when the polarity is held constant. Keeping polarity fixed, we count how many distinct answers the three prompt templates with slightly different wording the model yields. As we use three different prompt templates per polarity, there are three distinct values that can arise form this: full agreement, 2-vs-1 or complete disagreement. The results are visualized in \Cref{fig:app_tmr_detail} per model and dataset.
% Findings
Complete disagreement (three different answers) is extremely rare ($<1 \%$ for all models). Most entity pairs fall in the green “full agreement” slice. The orange 2-vs-1 splits account for the rest. Generally, the bigger the model, the less it disagrees with itself based on prompt wording.

\subsection{Summary}
\label{sec:prompt-sensitivity-summary}

Across the three tested dimensions of prompt sensitivity, a pattern emerges.
Larger models show greater stability.
They produce tightly clustered numerical estimates, maintain agreement across prompt templates, and display little polarity bias.
Smaller models, in contrast, are substantially more affected by re-phrasings, with higher variability in numbers, lower consensus across prompt templates, and stronger polarity asymmetries.
These findings suggest that robustness to prompt wording strengthens with scale, though sensitivities remain across all model sizes.
\section{Error Analysis of the Extracted Numerical Attributes}
\label{sec:numex_error}

\begin{sidewaysfigure*}[htbp]
    \centering
    \includegraphics[width=\linewidth]{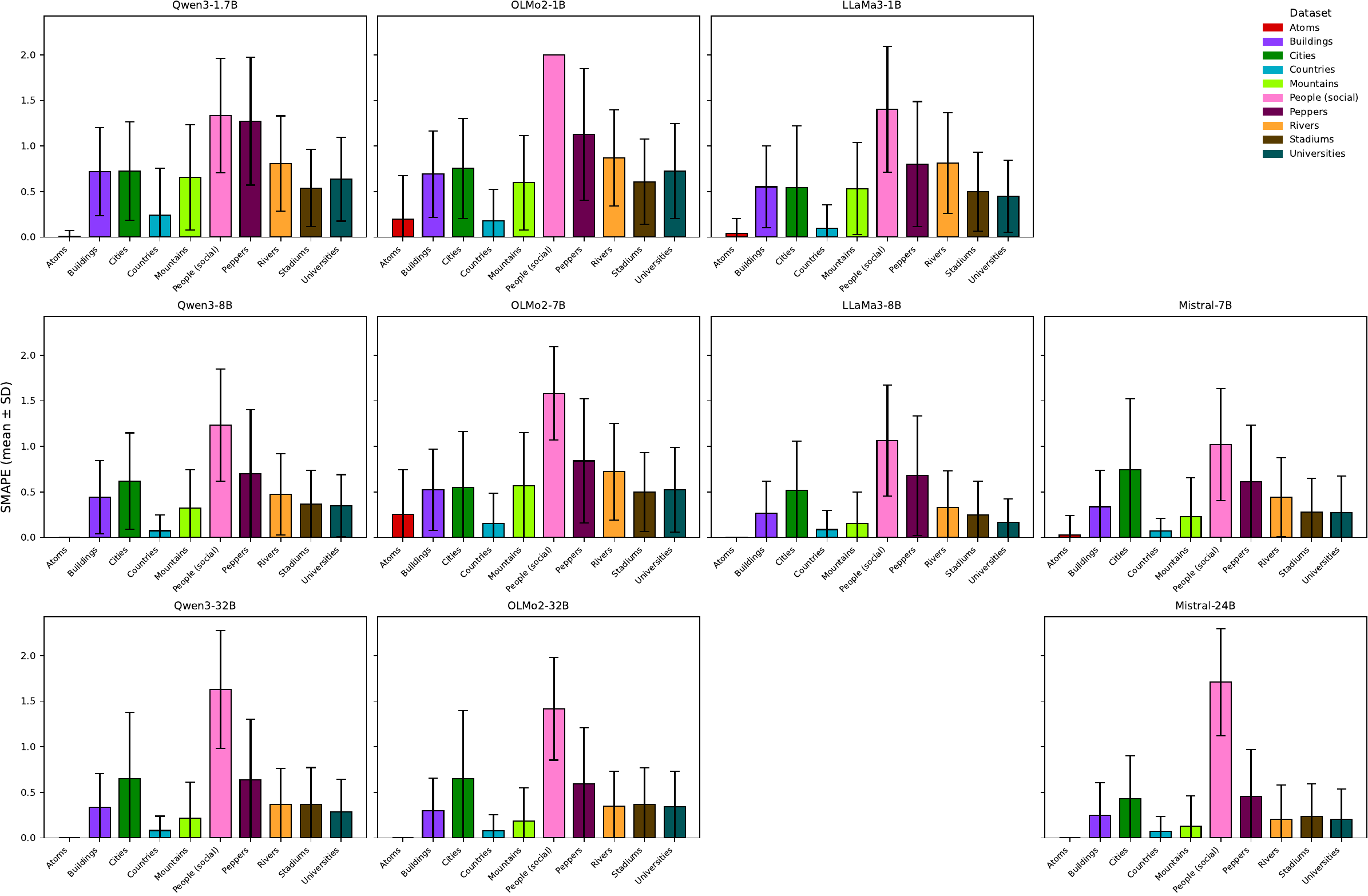}
    \caption{
        SMAPE ($\text{mean} \pm \text{SD}$) for each model–dataset pair.
        Bars show the mean SMAPE across extractions, with whiskers denoting
        standard deviation across samples. Lower values indicate more accurate
        numerical predictions.
    }
    \label{fig:app:numex_smape}
\end{sidewaysfigure*}

The main text is concerned with how well the numerical attributes extracted from the models can be used to rank entities. While this is the main goal of our study, it is also interesting to see how accurate the models are in predicting the actual numbers. As our analysis spans many numerical attributes of vastly different scales (e.g., lengths of rivers in the order of thousands of kilometers vs. number of social media followers in the order of millions), we need a scale-independent error metric to compare performance across datasets. Therefore, we use the Symmetric Mean Absolute Percentage Error (SMAPE),
\[
    \text{SMAPE}(y,\hat y)=
    \frac{|\hat y-y|}{(|\hat y|+|y|)/2},
\]
where $y$ is the ground-truth value and $\hat y$ the model's prediction. It is scale-independent and bounded in the interval $[0,2]$, making values directly comparable across quantities that span several orders of magnitude. A value of $0$ means perfect prediction, while $2.0$ represents the worst case.

\Cref{fig:app:numex_smape} shows the mean SMAPE for every model and dataset together with the standard deviation. The results reveal a mild size trend: \emph{larger models tend to achieve lower SMAPE scores}, meaning their extracted numbers are generally closer to the ground truth. However, even the largest models still make significant errors. A closer look into the per dataset results reveals that this can largely be attributed to a few datasets. All models perform best on the \texttt{atoms} dataset and worst on \texttt{people (social)}, with the remaining datasets showing similar values across architectures.
The \texttt{people (social)} dataset requires models to estimate the number of followers across various social media platforms, which can fluctuate significantly over time and is inherently difficult to approximate.
For datasets that are more stable, such as river lengths or mountain heights, the SMAPE is generally lower, while still not perfect even for the largest models.
SMAPE tells us what the model knows about the numbers, but not how useful this knowledge is in ranking tasks nor whether the model uses it.
These questions are explored in the main paper (see \Cref{sec:gener-model-perf}).

\section{Model Details}
\label{sec:app:model_info}

Unless specified otherwise, all models were run with greedy decoding and thinking was disabled, if applicable. Models with more than $10$B parameters were run in $8$ bit quantization.
All other models were run with $16$ bit floating point precision. 
An overview of all models used, along with citations and Hugging Face repository links, is provided in Table~\ref{tab:model_info}.

\begin{table*}[t]
    \footnotesize
    \centering    
    \begin{tabular}{ll}
        \toprule
        \textbf{Model} & \textbf{Hugging Face Repository} \\
        \midrule
        LLaMa3-1B \cite{grattafiori_llama_2024}        & meta-llama/Llama-3.2-1B-Instruct \\
        OLMo2-1B \cite{DBLP:journals/corr/abs-2501-00656}           & allenai/OLMo-2-0425-1B-Instruct \\
        Qwen3-1.7B \cite{qwen3}                        & Qwen/Qwen3-1.7B \\
        Mistral-7B \cite{DBLP:journals/corr/abs-2310-06825}           & mistralai/Mistral-7B-Instruct-v0.3 \\
        OLMo2-7B \cite{DBLP:journals/corr/abs-2501-00656}            & allenai/OLMo-2-1124-7B-Instruct \\
        LLaMa3-8B \cite{grattafiori_llama_2024}        & meta-llama/Llama-3.1-8B-Instruct \\
        Qwen3-8B \cite{qwen3}                          & Qwen/Qwen3-8B \\
        Mistral-24B \cite{mistral_small_3}             & mistralai/Mistral-Small-24B-Instruct-2501 \\
        OLMo2-32B \cite{DBLP:journals/corr/abs-2501-00656}        & allenai/OLMo-2-0325-32B-Instruct \\
        Qwen3-32B \cite{qwen3}                         & Qwen/Qwen3-32B \\
        \bottomrule
    \end{tabular}
    \caption{Information about the models used in this paper.}
    \label{tab:model_info}    
\end{table*}
\section{Detailed Accuracy Analysis}
\label{sec:app:detailed_acc}

\begin{sidewaysfigure*}[htbp]
    \includegraphics[width=\linewidth]{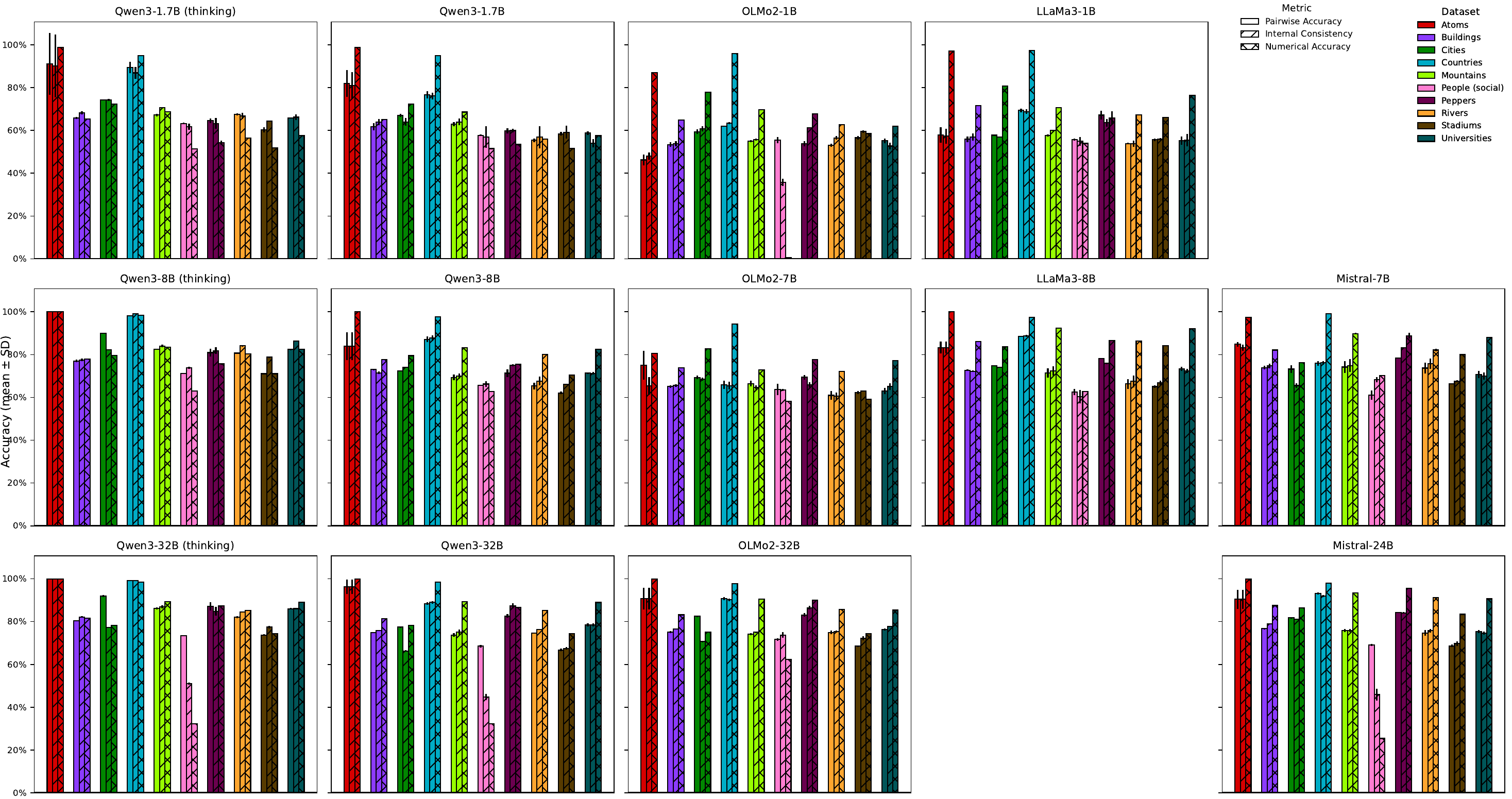}
    \caption{\textbf{Accuracy comparison for positive polarity prompts}. Each panel shows results for a single language model. For each dataset, we display three side-by-side bars: the solid bar represents the
        ranking accuracy, the hatched bar shows the internal consistency and the cross-hatched bars show the numerical accuracy. Bars indicate mean accuracy; error bars show $\pm$1 standard
        deviation across prompt templates.}
    \label{fig:acc_pos_pol_detailed}
\end{sidewaysfigure*}
\begin{sidewaysfigure*}[htbp]
    \includegraphics[width=\linewidth]{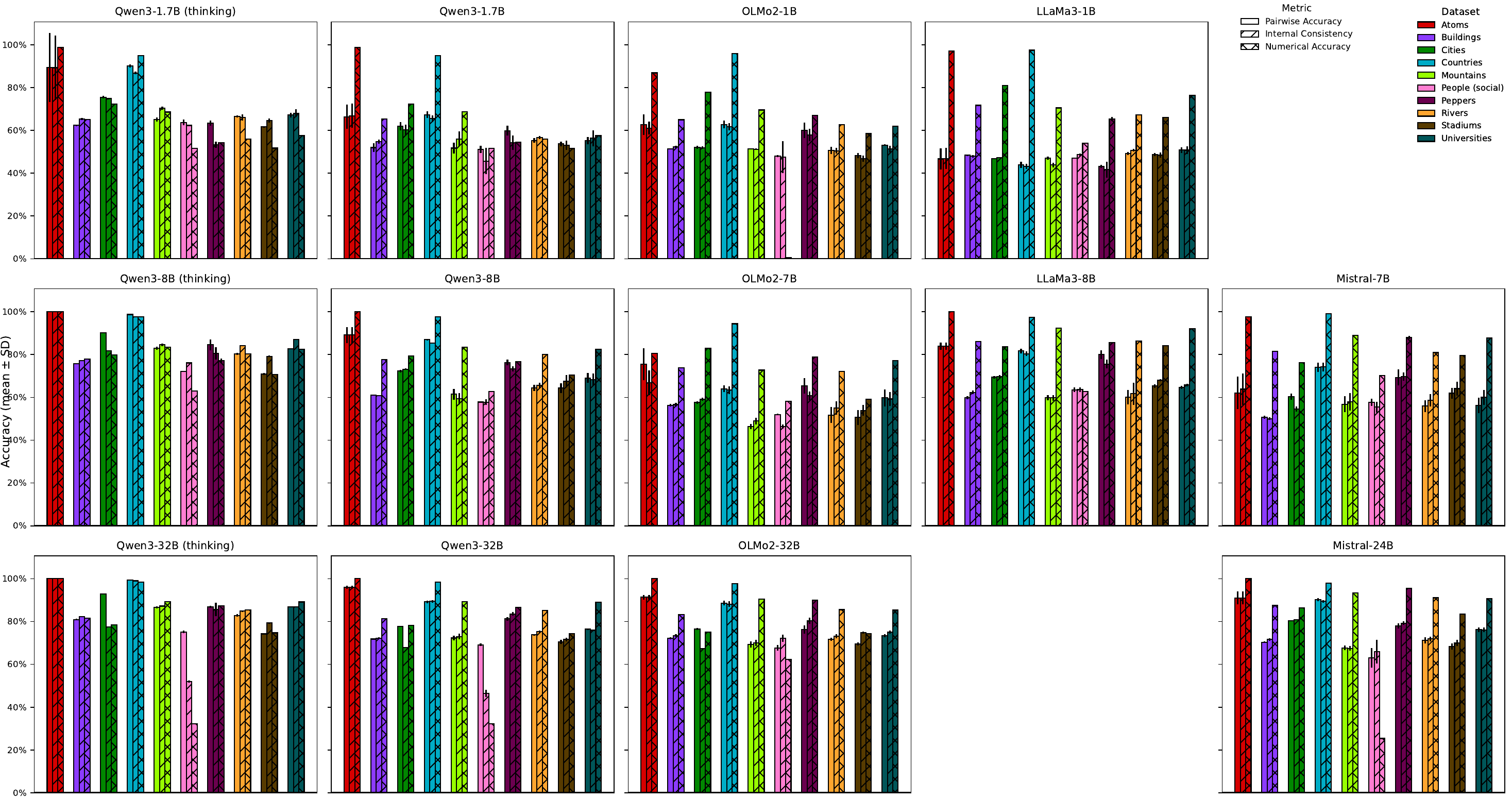}
    \caption{\textbf{Accuracy comparison for negative polarity prompts}. Same layout as \Cref{fig:acc_pos_pol_detailed}, but for negative polarity prompts.}
    \label{fig:acc_neg_pol_detailed}
\end{sidewaysfigure*}

As there are some differences in the results between prompts with positive and negative polarity, we report results for these types of prompts separately. 
\Cref{fig:acc_pos_pol_detailed,fig:acc_neg_pol_detailed} report accuracy comparisons for ranking accuracy, internal consistency, and numerical accuracy under positive and negative polarity prompts, respectively.
Each panel corresponds to a specific model, and each group of bars represents performance on one dataset.

\section{Co-occurrence Details}
\label{appendixDetailsCooccurrenceScores}

Table~\ref{tab:pmi_examples} provides qualitative examples in which cosine similarity to attribute-related keywords (e.g., “larger,” “bigger,” “more”) suggests the wrong ranking, which would lead models that rely on co-occurrence bias to make an incorrect prediction. The list of positive and negative keywords used to construct the “bigger-smaller” axis is shown in Table~\ref{tab:kewword_info}.

\begin{table*}[p]
    \centering
    \footnotesize
    \begin{tabular}{lll}
        \toprule
        \textbf{Dataset} & \textbf{Cosine Suggests}                    & \textbf{Actually Larger}                          \\
        \midrule
        People (social)  & George Michael ($\sim559$k followers)       & Mackenyu ($\sim1$M followers)                     \\
        Buildings        & Red Fort (33 m)                             & Colonius (266 m)                                  \\
        Atoms            & chromium (24)                               & niobium (41)                                      \\
        Universities     & University of Mannheim ($\sim12$k students) & George Washington University ($\sim24$k students) \\
        Peppers          & jalapeño (20k SHU)                          & Pepper X (3.1M SHU)                               \\
        Cities           & Palermo ($\sim674$k inhabitants)            & Islamabad ($\sim1.9$M inhabitants)                \\
        Stadiums         & Bolt Arena ($\sim10$k capacity)             & Kashima Stadium ($\sim40$k capacity)              \\
        Countries        & Botswana ($\sim2.4$M inhabitants)           & Yemen ($\sim2.8$M inhabitants)                    \\
        Mountains        & Mount Scenery (887 m)                       & Half Dome (2693 m)                                \\
        Rivers           & Mystic River (113 km)                       & Bega River (256 km)                               \\
        \bottomrule
    \end{tabular}
    \caption{Examples of entity pairs where similarity to the considered keywords disagrees with the ground truth numerical values. The first entity is the one with the higher cosine similarity to the keywords, but with a lower numerical value.}
    \label{tab:pmi_examples}
\end{table*}

\begin{table*}[p]
    \centering
    \footnotesize
    \begin{tabular}{lll}
        \toprule
        \textbf{Dataset} & \textbf{Positive keywords}                   & \textbf{Negative keywords}                \\
        \midrule
        Atoms            & heaviest, largest, highest, massive, big     & lightest, smallest, lowest, tiny, low     \\
        Buildings        & tallest, highest, largest, big, tall         & shortest, smallest, lowest, tiny, low     \\
        Cities           & largest, populous, big, crowded, dense       & smallest, quiet, tiny, remote, sparse     \\
        Countries        & largest, populous, big, powerful, dense      & smallest, sparse, tiny, quiet, remote     \\
        Mountains        & highest, tallest, largest, elevated, big     & lowest, smallest, shortest, low, tiny     \\
        Peppers          & hottest, spiciest, pungent, intense, fiery   & mildest, bland, cool, weak, low           \\
        People (birth)   & youngest, recent, modern, newer, late        & oldest, ancient, early, historic, vintage \\
        People (social)  & popular, famous, followed, liked, viral      & unknown, obscure, ignored, unseen, small  \\
        Rivers           & longest, largest, broadest, deep, big        & shortest, smallest, shallow, narrow, tiny \\
        Stadiums         & largest, busiest, crowded, massive, big      & smallest, quiet, empty, tiny, low         \\
        Universities     & largest, populous, crowded, big, prestigious & smallest, quiet, tiny, local, low         \\
        \bottomrule
    \end{tabular}
    \caption{List of positive and negative keywords that are used to capture co-occurrence bias. The positive keywords are terms that are associated with high values of the considered attribute, negative keywords are associated with low values.}
    \label{tab:kewword_info}
\end{table*}
\section{Bias Alignment with Ground Truth}
\label{sec:bias_gt_align}

\begin{table*}[p]
    \scriptsize
    \begin{tabular}{lcccccccccc}
        \toprule
        Dataset       & Atoms & Buildings & Cities & Countries & Mountains & People (social) & Peppers & Rivers & Stadiums & Universities \\
        \midrule
        Popularity    & 24\%  & 50\%      & 63\%   & 68\%      & 56\%      & 58\%            & 61\%    & 61\%   & 65\%     & 61\%         \\
        Co-occurrence & 50\%  & 55\%      & 61\%   & 68\%      & 52\%      & 52\%            & 58\%    & 54\%   & 51\%     & 55\%         \\
        \bottomrule
    \end{tabular}
    \caption{Cue--ground-truth alignment per dataset. Note that this will be the same for all models, as all models were evaluated on the same pairs.}
    \label{tab:bias_alignment}
\end{table*}

Table~\ref{tab:bias_alignment} shows that for most datasets, popularity points to the larger item substantially more than half of the time.
One outlier is the Atoms dataset, where popularity corresponds to the largest items only 24\% of the time.
Lighter elements tend to be more common (such as hydrogen, carbon, or oxygen) and thus more popular.
We also find that the co‐occurrence cue lines up with the ground truth more than half of the time.
Mentions of entities such as the Mount Everest or the Nile River tend to co-occur with adjectives expressing their magnitude, because they are the largest or longest in their respective categories.
\section{BOS details}

\subsection{E-Values}
\label{sec:bias_e_value}

\begin{figure*}[t]
    \includegraphics[width=\linewidth]{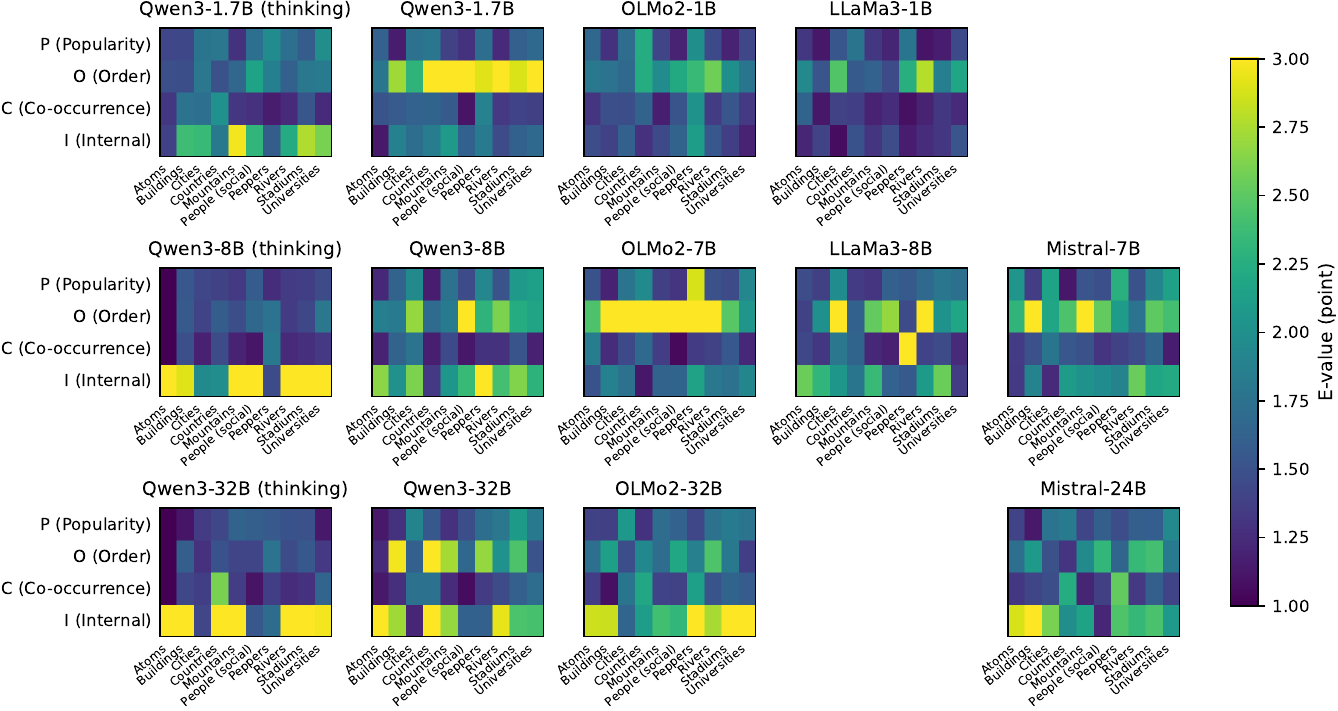}
    \caption{
        E-values, which quantify the strength of an unmeasured confounder (on the RR scale) needed to fully explain away the observed effect. Higher values indicate more robust evidence for a causal effect.
    }
    \label{fig:evalues}
\end{figure*}

While the BOS makes the measured cues independent of each other, it does not remove bias from \emph{unmeasured} confounders.
To assess how robust our results are to such confounding, we compute the E-value \citep{vanderweele_sensitivity_2017} for each estimated RR with $\text{RR} \ge 1$:
\[
    \text{E-Value} \;=\; \text{RR} + \sqrt{\text{RR} \times (\text{RR} - 1)}
\]
Larger E-values indicate more robust evidence that the observed effect is not solely due to unmeasured confounding.
It can be interpreted as the minimum strength of association (on the RR scale) that an unmeasured confounder would need to have with both the cue and the outcome in order to fully account for the observed effect.
For instance, an E-value of $1.5$ means that a hidden confounder would need to be associated with both the cue and the outcome by a risk ratio of at least $1.5$ to nullify the effect.

The E-values in Figure \ref{fig:evalues} show that order cues yield the highest robustness scores, especially for smaller models (e.g., $\text{E-value} = 3.28$ for Qwen3-1.7B).
Popularity and co-occurrence have lower E-values ($\approx 1.1-1.6$), suggesting they could be more easily explained by unmeasured variables, while internal alignment effects in larger models show E-values above $2$, indicating comparatively strong causal robustness.

\subsection{BOS Discard Info}
\label{app:sec:bos_discard}
The BOS balancing procedure necessarily discards any surplus items beyond the least-frequent (P, T, C, I) combination in each template.
\crefrange{tab:bos-discard-part 1}{tab:bos-discard-part 2} detail the retained‐over‐total counts per dataset and model.
Although the absolute number of discarded items can be sizeable, enough balanced examples remain in every dataset–model pair to yield narrow confidence intervals in Figure \ref{fig:biases}, confirming that the subsequent cue-ablation results are not affected by data scarcity.

\begin{table*}[htbp]
    \centering
    \footnotesize
    \begin{tabular}{lccccc}
        \toprule
        Model           & LLaMa3-1B     & LLaMa3-8B     & Mistral-24B  & Mistral-7B   & OLMo2-1B      \\
        \midrule
        Atoms           & $90/2812$     & $1200/2831$   & $1200/2832$  & $81/2653$    & $384/2828$    \\
        Buildings       & $7664/20844$  & $6336/21129$  & $4512/21068$ & $6064/20133$ & $10464/21276$ \\
        Cities          & $6336/22860$  & $6672/22870$  & $4992/22872$ & $5264/21722$ & $6512/22844$  \\
        Countries       & $90/4703$     & $96/4699$     & $90/4703$    & $77/4625$    & $96/4700$     \\
        Mountains       & $6976/14437$  & $2160/14642$  & $1440/15051$ & $2096/12755$ & $8240/14792$  \\
        People (birth)  & $28400/88796$ & $5312/88457$  & $2352/88393$ & $3456/78493$ & $22080/88607$ \\
        People (social) & $9840/18922$  & $10176/19404$ & $1440/19176$ & $5360/19305$ & $84/19152$    \\
        Peppers         & $144/714$     & $78/864$      & $60/828$     & $90/791$     & $192/862$     \\
        Rivers          & $6416/17933$  & $4224/18237$  & $2096/18403$ & $4784/17159$ & $9888/18246$  \\
        Stadiums        & $9312/18691$  & $5712/18472$  & $5280/18455$ & $5184/17690$ & $10752/18603$ \\
        Universities    & $8064/23608$  & $8784/23647$  & $2832/23649$ & $3552/22658$ & $8160/23539$  \\
        \bottomrule
    \end{tabular}
    \caption{Retained-over-total counts per dataset and
        model (part 1)}
    \label{tab:bos-discard-part 1}
\end{table*}

\begin{table*}
    \centering
    \footnotesize
    \begin{tabular}{lccccc}
        \toprule
        Model           & OLMo2-32B    & OLMo2-7B      & Qwen3-1.7B    & Qwen3-32B    & Qwen3-8B     \\
        \midrule
        Atoms           & $1200/2832$  & $96/2832$     & $78/2831$     & $1200/2832$  & $1200/2832$  \\
        Buildings       & $6384/21270$ & $9312/21090$  & $10560/21120$ & $6912/21258$ & $8496/21240$ \\
        Cities          & $7200/22834$ & $6528/22870$  & $7680/22835$  & $4800/22872$ & $7008/22815$ \\
        Countries       & $96/4704$    & $288/4668$    & $192/4703$    & $90/4704$    & $90/4449$    \\
        Mountains       & $2640/14976$ & $5760/14748$  & $7680/14819$  & $2544/14901$ & $3888/15132$ \\
        People (birth)  & $5136/88982$ & $28608/88882$ & $18288/88598$ & $5616/88998$ & $9120/88571$ \\
        People (social) & $9360/19428$ & $10272/19020$ & $8352/19234$  & $4368/19247$ & $8016/18983$ \\
        Peppers         & $78/828$     & $96/864$      & $288/827$     & $78/864$     & $90/828$     \\
        Rivers          & $4032/18643$ & $8640/18225$  & $7728/18302$  & $3872/18322$ & $5952/18351$ \\
        Stadiums        & $7392/18525$ & $9296/18589$  & $8352/18518$  & $7200/18356$ & $7104/18325$ \\
        Universities    & $4896/23652$ & $7584/23499$  & $9312/23628$  & $4128/23709$ & $5136/23712$ \\
        \bottomrule
    \end{tabular}
    \caption{Retained-over-total counts per dataset and
        model (part 2)}
    \label{tab:bos-discard-part 2}
\end{table*}

\subsection{Per-Dataset Cue Effects}
\label{app:sec:bos_detailed}
\begin{sidewaysfigure*}[htbp]
    \includegraphics[width=\linewidth]{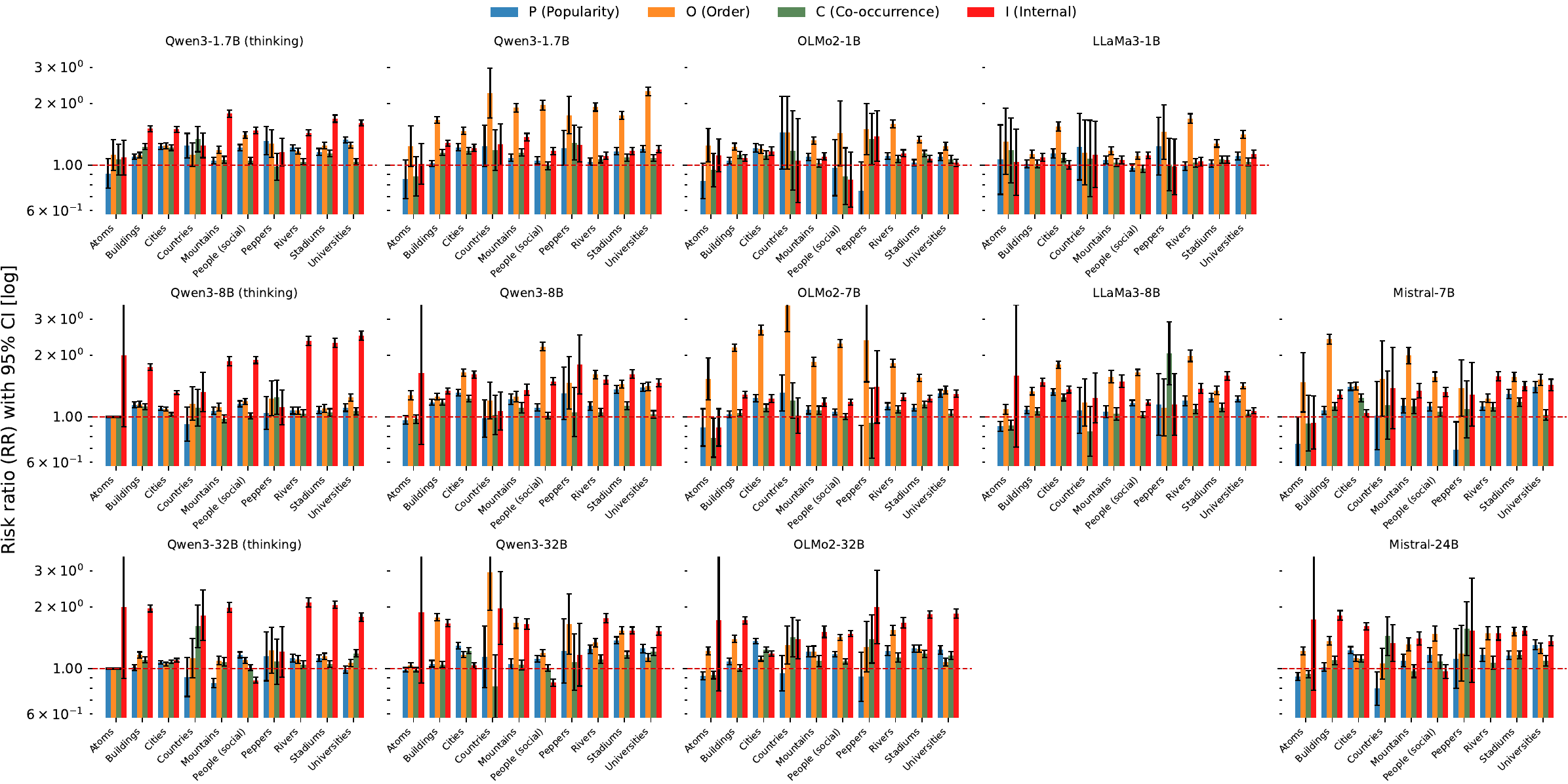}
    \caption{Per-dataset, per-model feature ablation analysis on the Balanced–Orthogonal Subset (BOS).\@
        Each block of four bars reports the accuracy change that follows when a single feature is flipped while all other observable factors are fixed:\@
        popularity \((P)\), mention order \((O)\), cosine similarity \((C)\), and internal–ground-truth alignment \((I)\).\@
        Signed gaps are shown for \(P\), \(C\), and \(I\); an absolute gap is shown for \(O\) because its direction can reverse with prompt polarity or model family.\@
        Whiskers indicate $\pm1$ standard deviation across prompt templates.}
    \label{fig:bos_detailed}
\end{sidewaysfigure*}

For completeness, Figure \ref{fig:bos_detailed} breaks down the BOS cue-ablation gaps by dataset and model, complementing the aggregate view in Figure \ref{fig:biases}.

\subsection{Risk Ratios}
\label{app:risk-ratio}

The BOS risk ratios are within-model quantities and depend on the model’s baseline accuracy when a cue is absent. Algebraically, $\text{RR} = 1 + \Delta/p_0$, where $p_0 = \Pr\!\bigl(Y=\hat{Y}\mid F=0\bigr)$ and $\Delta$ is the absolute accuracy lift when the cue is present, i.e., $\Pr\!\bigl(Y=\hat{Y}\mid F=1\bigr) - \Pr\!\bigl(Y=\hat{Y}\mid F=0\bigr)$. Thus, the same $\Delta$ yields a larger RR for a low-baseline model and a smaller RR for a high-baseline model. The effect is especially pronounced for $I$ (number-GT alignment), because $I$ is based on the model's extracted numbers. Consequently, the sets with $I=0$ and $I=1$ differ across models, leading to vastly different baselines $p_0$. In addition, noise in the extracted numbers \cite{allen-zhu_physics_2024-1,schwartz-etal-2024-numerologic, DBLP:conf/iclr/YangHKLZ25} might mislabel some $I$ cases, attenuating $RR_I$ towards $1$. A model can therefore rely heavily on its numbers yet show a modest $RR_I$, simply because it already performs well when $I=0$ (large $p_0$). Therefore, $RR$ values are best interpreted in an intra-model fashion.
\section{Detailed Meta-predictor Results}
\label{sec:detailed_meta_pred}

\Cref{tab:meta-pred-acc} shows the accuracies of the bias-only meta-predictor broken down by dataset and model.

\begin{table*}[t]
    \centering
    \scriptsize
\setlength{\tabcolsep}{2pt}

\begin{tabular}{lcccccccccc}
\toprule
Model & LLaMa3-1B & OLMo2-1B & Qwen3-1.7B & Mistral-7B & OLMo2-7B & LLaMa3-8B & Qwen3-8B & Mistral-24B & OLMo2-32B & Qwen3-32B \\
Dataset &  &  &  &  &  &  &  &  &  &  \\
\midrule
Atoms & $53\% \pm 5\%$ & $65\% \pm 6\%$ & $66\% \pm 6\%$ & $69\% \pm 10\%$ & $72\% \pm 5\%$ & $72\% \pm 6\%$ & $72\% \pm 4\%$ & $74\% \pm 5\%$ & $75\% \pm 4\%$ & $75\% \pm 3\%$ \\
Buildings & $57\% \pm 2\%$ & $57\% \pm 2\%$ & $66\% \pm 6\%$ & $73\% \pm 6\%$ & $71\% \pm 5\%$ & $59\% \pm 3\%$ & $61\% \pm 4\%$ & $58\% \pm 3\%$ & $61\% \pm 5\%$ & $64\% \pm 3\%$ \\
Cities & $60\% \pm 2\%$ & $58\% \pm 4\%$ & $66\% \pm 4\%$ & $69\% \pm 3\%$ & $75\% \pm 5\%$ & $72\% \pm 1\%$ & $69\% \pm 2\%$ & $66\% \pm 2\%$ & $68\% \pm 2\%$ & $68\% \pm 2\%$ \\
Countries & $62\% \pm 5\%$ & $63\% \pm 5\%$ & $73\% \pm 5\%$ & $75\% \pm 3\%$ & $81\% \pm 4\%$ & $69\% \pm 6\%$ & $70\% \pm 3\%$ & $69\% \pm 4\%$ & $72\% \pm 3\%$ & $70\% \pm 4\%$ \\
Mountains & $56\% \pm 2\%$ & $57\% \pm 4\%$ & $68\% \pm 7\%$ & $69\% \pm 7\%$ & $65\% \pm 5\%$ & $65\% \pm 4\%$ & $61\% \pm 6\%$ & $61\% \pm 5\%$ & $59\% \pm 3\%$ & $64\% \pm 3\%$ \\
People (social) & $55\% \pm 2\%$ & $60\% \pm 5\%$ & $74\% \pm 14\%$ & $64\% \pm 6\%$ & $80\% \pm 12\%$ & $69\% \pm 8\%$ & $71\% \pm 8\%$ & $65\% \pm 7\%$ & $65\% \pm 6\%$ & $60\% \pm 2\%$ \\
Peppers & $62\% \pm 8\%$ & $63\% \pm 8\%$ & $65\% \pm 8\%$ & $63\% \pm 5\%$ & $63\% \pm 7\%$ & $59\% \pm 9\%$ & $60\% \pm 5\%$ & $59\% \pm 10\%$ & $58\% \pm 11\%$ & $57\% \pm 9\%$ \\
Rivers & $64\% \pm 8\%$ & $62\% \pm 5\%$ & $70\% \pm 10\%$ & $60\% \pm 4\%$ & $67\% \pm 9\%$ & $69\% \pm 7\%$ & $63\% \pm 4\%$ & $64\% \pm 4\%$ & $64\% \pm 2\%$ & $65\% \pm 2\%$ \\
Stadiums & $57\% \pm 2\%$ & $57\% \pm 5\%$ & $65\% \pm 4\%$ & $66\% \pm 5\%$ & $62\% \pm 6\%$ & $66\% \pm 2\%$ & $66\% \pm 2\%$ & $64\% \pm 3\%$ & $65\% \pm 2\%$ & $69\% \pm 2\%$ \\
Universities & $60\% \pm 5\%$ & $59\% \pm 4\%$ & $74\% \pm 8\%$ & $64\% \pm 2\%$ & $62\% \pm 4\%$ & $64\% \pm 4\%$ & $64\% \pm 2\%$ & $63\% \pm 3\%$ & $61\% \pm 4\%$ & $62\% \pm 3\%$ \\
\midrule
Avg. & $59\% \pm 4\%$ & $60\% \pm 5\%$ & $69\% \pm 7\%$ & $67\% \pm 5\%$ & $70\% \pm 6\%$ & $66\% \pm 5\%$ & $66\% \pm 4\%$ & $64\% \pm 5\%$ & $65\% \pm 4\%$ & $65\% \pm 3\%$ \\
\bottomrule
\end{tabular}
% end resize
\caption{Bias-only meta-predictor accuracy (\% mean $\pm$ std) by dataset (rows) and model (columns).
    The meta-predictor is a logistic regression that uses only surface cues: QRank popularity, positional advantage, and cosine association with “bigger’’ terms.
    Values are averaged over prompt templates with $5$-fold cross-validation; the bottom row reports the macro-average across datasets for each model. Higher is better.}
\label{tab:meta-pred-acc}
\end{table*}
\section{Detailed Case Analyses}
\label{sec:detailed_cases}

\subsection{Per Dataset Dataset Distribution of Cases}
\label{app:sec:cases_per_dataset}
\begin{sidewaysfigure*}[htbp]
    \includegraphics[width=\linewidth]{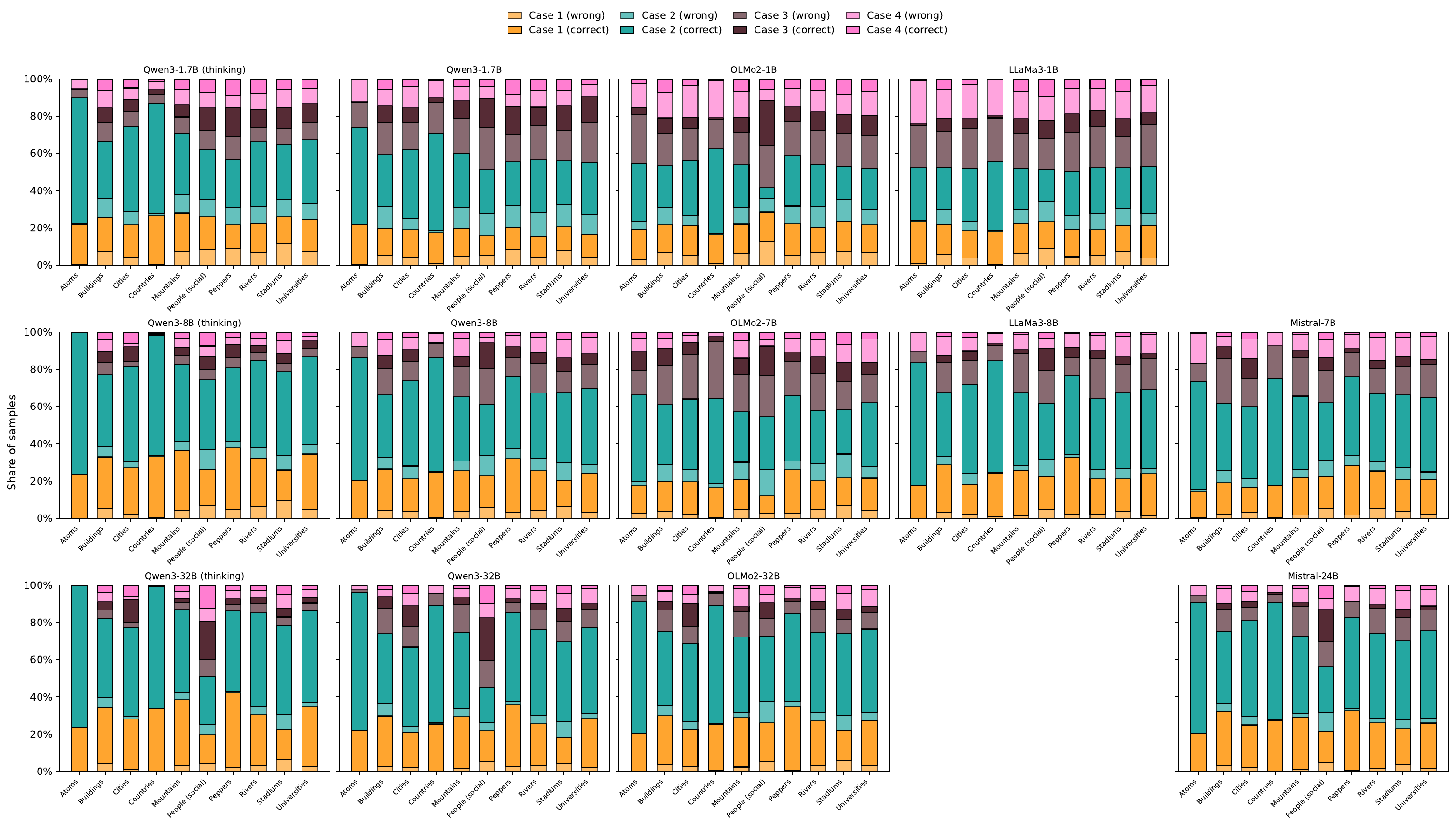}
    \caption{
        Detailed breakdown of the four cases (from Section \ref{sec:meta-predictor}) per dataset and model.
        Within every panel each colored, stacked bar corresponds to one dataset.
        The segments show the proportion of test samples that fall into
        \textcolor{Case1Color}{\textbf{Case 1}} (numerical and pairwise agree, meta-predictor disagrees),
        \textcolor{Case2Color}{\textbf{Case 2}} (all three agree),
        \textcolor{Case3Color}{\textbf{Case 3}} (pairwise aligns with surface-bias meta-predictor, but not with numerical comparison), and
        \textcolor{Case4Color}{\textbf{Case 4}} (neither numbers nor meta-predictor explain the choice).
        This figure complements \Cref{fig:case_distribution} by revealing \emph{which datasets drive each pattern} for every model.  Percentages are computed after filtering out samples with unanswerable prompts or missing extractions; bars sum to 100\,\% within each dataset. Opaque segments marks correct predictions and translucent segment mark errors with respect to the ground truth.}
    \label{fig:meta_pred_cases_detailed}
\end{sidewaysfigure*}

This section provides a deeper analysis of the two meta-predictors introduced in Section~\ref{sec:meta-predictor}.
Figure~\ref{fig:meta_pred_cases_detailed} presents a breakdown of the four diagnostic cases discussed in Section~\ref{sec:meta-predictor}.
Each panel corresponds to a different model, and each bar to a dataset. This figure complements the main paper’s analysis by revealing which types of errors are most prevalent in each domain, and whether failures to follow numerical predictions correlate with surface-level biases.

\subsection{Case 2 Detailed Analysis}
\label{app:sec:case2_swap}

\begin{sidewaysfigure*}[htbp]
    \includegraphics[width=\linewidth]{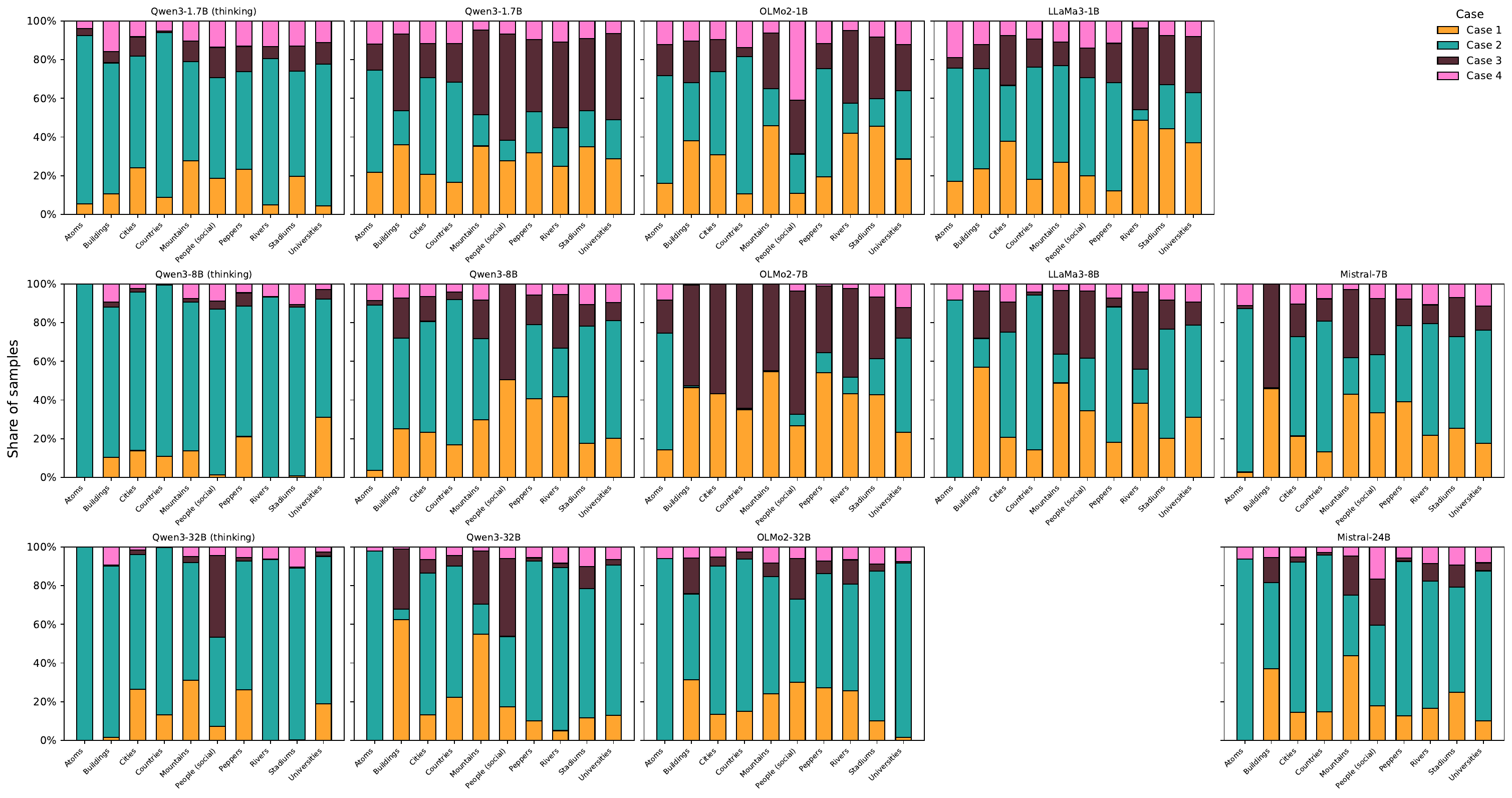}
    \caption{Per-dataset, per-model breakdown of the mention-order swap
        experiment.
        Bars show how items that were originally in Case 2 migrate to Cases 1–4
        after reversing the entity order.}
    \label{app:fig:case2}
\end{sidewaysfigure*}

In the original taxonomy in \Cref{sec:cases}, Case~2 gathers all samples for which the model's own numerical comparison and the meta-predictor point to the same answer. Because the signals are perfectly aligned, we cannot tell which one actually drives the decision. Luckily, mention order is the strongest single bias we have identified (\Cref{fig:biases}) and trivial to reverse without altering anything else. We therefore reran each Case~2 prompt with the two entities swapped, keeping every other token unchanged. The swap leaves popularity and cosine cues untouched but inverts the positional feature. After that we, re-classify the samples into the four cases. An interpretation of what it means for a sample to end up in each case after having its order swapped is summarized in \cref{tab:case2_swap_interpretation}.

\begin{table*}
    \footnotesize
    \begin{tabularx}{\linewidth}{lXX}
        \toprule
        \textbf{New Case} & \textbf{Observation}                                                      & \textbf{Interpretation}                                                                             \\
        \midrule
        Case~1            & Model keeps its original choice; meta-predictor now expects the opposite. &
        Decision is anchored in internal numerical knowledge; positional cue was not decisive.      
        \\
        Case~2            & Model and meta-predictor both remain unchanged.                           & Signals still coincide. 
        We cannot disentangle whether numbers or biases drove the choice.                                                                                                                                   \\
        Case~3            & Model changes its answer exactly as the meta-predictor predicts.          & Positional cue overrides numerical preference; behavior is dominated by surface heuristics.         \\
        Case~4            & Model changes its answer to an outcome neither explanation predicts.      & Residual noise, reliance on unmodeled cues or misjudgment by meta-predictor; indicates instability. \\
        \bottomrule
    \end{tabularx}
    \caption{Interpretation of the four cases after swapping the entity order in Case~2 samples.\label{tab:case2_swap_interpretation}}
\end{table*}

In brief: if the model answers with the same entity while the meta-predictor disagrees (Case~1) it suggests that the original response was anchored in the model's own numbers. When both model and meta-predictor remain aligned (Case~2) the two signals are still inseparable. A change in the model's answer that the meta-predictor correctly anticipates (Case~3) betrays domination by the positional cue.
Finally, a change in the model's answer neither explanation anticipated (Case~4) points to residual noise or to biases that our simple meta-predictor does not capture.

% Results
\Cref{app:fig:case2} shows how the once-ambiguous Case-2 samples redistribute across the four cases after the order of the entities is swapped.
The case distribution after swapping reveals size-dependent patterns.
Smaller and mid-sized models (e.g., Qwen3-1.7B, OLMo2-7B) often transition Case 2 items into Case 3, showing that mention order alone can override internal reasoning.
OLMo2-7B is especially prone, which is consistent with the strong position bias observed in the BOS analysis (\cref{fig:biases}).
The smallest model we tested shows a notably high Case~4 rate, both before and after the swap (see \cref{fig:case_distribution}).
Neither the biases nor numerical reasoning explain these predictions well.
Their answers appear noisy rather than systematic.
Larger models (e.g., OLMo2-32B, Mistral-24B) show the opposite trend.
Roughly half of their Case 2 items remain unchanged after the swap, and about 30\% move into Case 1, indicating that their choices are anchored in numerical representations.
Case 4 remains rare.
Notably, the $\approx 30 \%$ transition into Case 1 is stable across almost all models, suggesting that once the numerical signal dominates, it does so reliably.
Note that this may reflect "easy" comparisons with large numerical gaps; we investigate this further in \Cref{sec:case1_vs_case3}.

\subsection{Case 1 vs Case 3 Details}
\label{app:sec:case1_vs_3}

\begin{sidewaysfigure*}[p]
    \includegraphics[width=\linewidth]{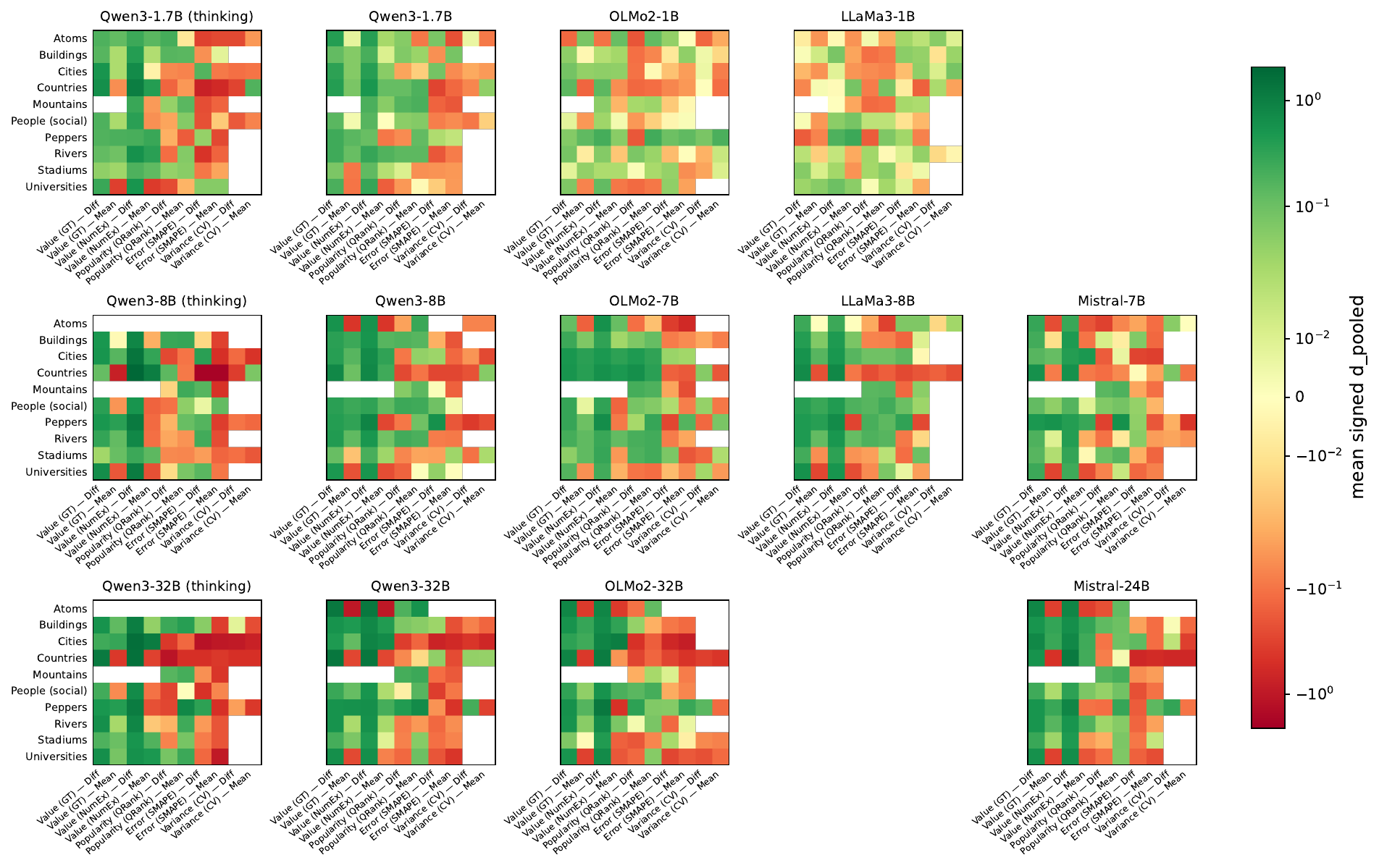}
    \caption{Per-model, per-dataset signed Cohen’s $d$ for the Case~1 vs Case~3 contrast, pooled over prompt polarity. Columns list the feature summaries used in the main figure (for each family: Value, Popularity, NumEx error, NumEx variance, showing Diff and Mean). Colors use a symmetric-log scale (green = larger in Case 1, red = larger in Case 3; shared colorbar across panels). White cells denote missing estimates due to insufficient support for one case or zero within-group variance}
    \label{app:fig:case1_vs_case3}
\end{sidewaysfigure*}

\Cref{app:fig:case1_vs_case3} breaks the Case 1 vs Case 3 effects down by model (panels) and dataset (rows) while keeping the feature set (columns) identical to the figure from the main paper. The per-dataset view largely mirrors the global pattern from \Cref{fig:case1_vs_case3}.
The strength of the observed effects varies by dataset and model size, with larger models typically showing clearer positive value gaps and more negative error/variance means. White cells indicate that $d$ could not be estimated for that model–dataset–feature triplet (e.g., one of the cells had no support after filtering to Cases 1/3 or the within-group variance collapsed to zero).

\section{Cohen's \texorpdfstring{$d$}{d}}
\label{sec:appendix_cohen}
We briefly recall how Cohen's $d$ is computed, applied to the specific setting where we want to compare feature values between Cases 1 and 3. For each feature $x$, we first split the data into the two groups and compute sample means and variances:
\begin{align*}
    \mu_{1}      & = \text{mean}(x \mid \text{Case 1}) \\
    \mu_{0}      & = \text{mean}(x \mid \text{Case 3}) \\
    \sigma_{1}^2 & = \text{var}(x \mid \text{Case 1})  \\
    \sigma_{0}^2 & = \text{var}(x \mid \text{Case 3}).
\end{align*}
We then form the pooled standard deviation, which summarizes the typical spread inside the two groups:
\begin{align*}
    \sigma_p^2 & =\tfrac12\!\left(\sigma_{1}^2+\sigma_{0}^2\right) \\
    \sigma_p   & =\sqrt{\sigma_p^2}
\end{align*}
Finally we standardize the mean difference
\[
    d \;=\; \frac{\mu_{1}-\mu_{0}}{\sigma_p}.
\]
\section{Detailed CoT Analysis}
\label{sec:app:detailed_cot}

\begin{figure*}[t]
        \includegraphics[width=\linewidth]{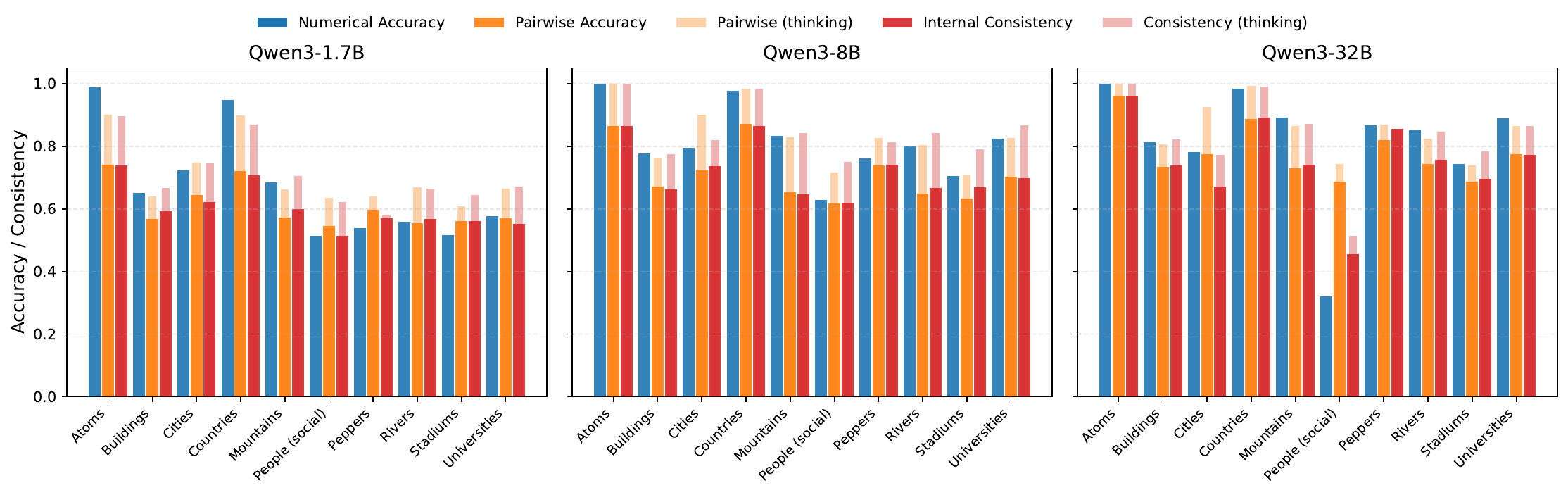}
        \caption{Comparison of the performance with and without thinking, in terms of pairwise accuracy and internal consistency. The figure also reports numerical accuracy (without thinking).}
        \label{fig:thinking_acc_detailed}
\end{figure*}

\begin{figure*}[t]
        \includegraphics[width=\linewidth]{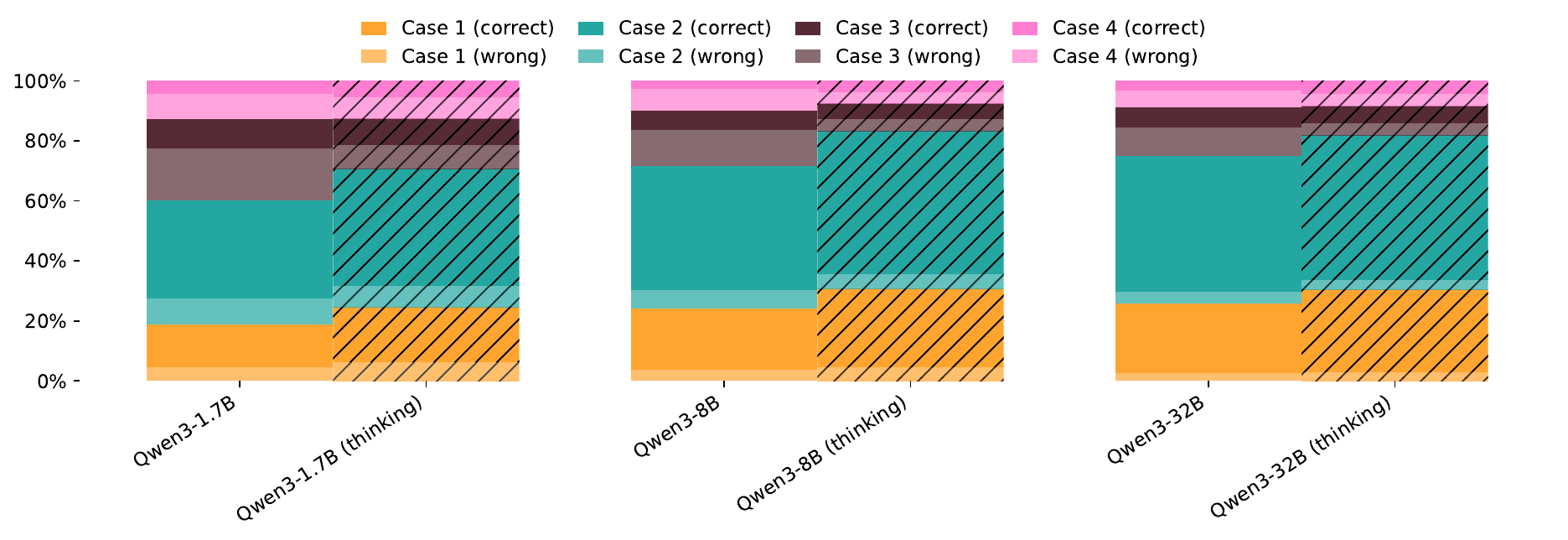}
        \caption{Case distribution with and without thinking.}
        \label{fig:thinking_case_distribution}
\end{figure*}

\begin{figure*}[t]
        \includegraphics[width=\linewidth]{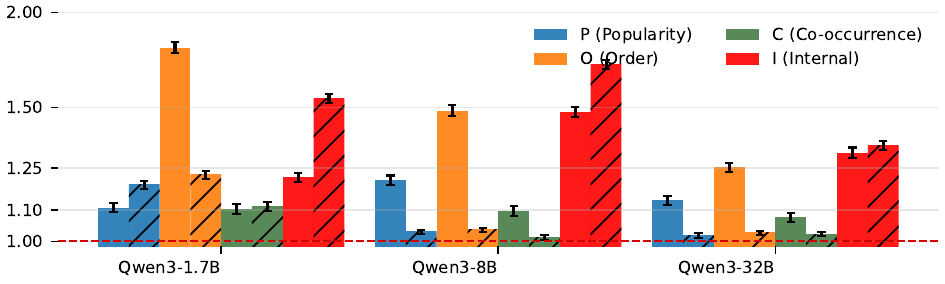}
        \caption{Risk ratios from BOS analysis with and without thinking.}
        \label{fig:thinking_bos}
\end{figure*}

% Results per dataset

\Cref{fig:thinking_acc_detailed}
compares the performance of Qwen3 models with and without thinking per dataset.

% Results case breakdown
% Meta Pred case analysis
Prompting models to ``think'' before answering changes the case distribution drastically.
% Case 1 and 2 rise
As \Cref{fig:thinking_case_distribution} shows, the largest gain comes from Case 1 ($\approx +6$ pp across models), followed by a smaller but still clear rise in Case 2 ($\approx + 5$ pp).
Taken together, these shifts mean that the final pairwise choice agrees more often with the model's own numerical comparison, as expected.

% Case 3 shrinks
The percentage of Case 3 items shrinks significantly, indicating a reduced tendency to follow surface-form cues when they conflict with the model's numerical comparison.
Among the Case 3 items that remain, the conditional correctness with respect to ground truth improves: if we look only at those remaining Case 3 rows, roughly half of them are correct with respect to the ground truth, whereas this share was lower without thinking.
Put plainly, when “thinking” models still side with non-numeric cues, those cues line up with ground truth more often than before. We hypothesize the model sometimes treats its numeric estimates as unreliable and gives partial weight to alternative and occasionally informative signals.
% Case 4 shrinks
Case 4 also shrinks overall, but the overall fraction of cases where it is right with respect to the ground-truth remains the same.
We thus find that explicit reasoning reduces bias-driven and noisy inconsistencies but does not achieve fully faithful numerical grounding.

% Link to acc and ic
This case redistribution aligns with our findings from \Cref{fig:thinking_acc}: the increase in pairwise accuracy is nearly mirrored by the increase in internal consistency.
That parallel movement has a simple cause: \emph{''thinking´´ mainly changes which signal the model follows at comparison time, not the quality of the signals themselves}.
The numerical extraction pipeline is unchanged, so the correctness of the numbers is fixed; what changes is that more decisions are made in agreement with those numbers, as seen in a shift of mass from Case~3/4 to Case~1/2.

% BIASES BOS RESULTS
As a causal cross-check, we repeat the BOS ablation with and without ``thinking'' (\Cref{fig:thinking_bos}). Bars above the dashed line indicate that the cue, when set to favor the correct answer, increases accuracy; hatched bars show the thinking variant. Across all three Qwen3 model sizes, reasoning clearly strengthens the Internal feature, meaning that \emph{decisions follow the model's own numeric comparison more often}. For the two largest models, all three of the surface cues decrease to negligible levels. For Qwen3-1.7B, the Popularity and Co-occurrence cues remain broadly similar; the  Order cue decreases in importance but remains decisive. As a reminder (see \Cref{app:risk-ratio}), BOS risk ratios are within-model quantities, hence bar heights should not be compared across models. In particular, the Internal bar appearing smaller for Qwen3-32B, when compared to the same bar for Qwen3-8B or Qwen3-1.7B in thinking mode, does not imply weaker numerical reliance.

\begin{table*}[t]
        \footnotesize
        \begin{tabularx}{\linewidth}{X}
                \toprule

                \textbf{Model:} \textsc{Qwen3-32B}
                \\[2pt]
                Mount Davis (predicted: $2160$m   actual: $979$m)                     \\[2pt]
                Grand Colombier (predicted: $1686$m   actual: $1534$m)                \\[2pt]
                \textbf{Trace:} Okay, let's see. The user is asking which mountain is lower between Mount
                Davis and Grand Colombier. I need to find the elevations of both.  First,
                Mount Davis. I'm not too familiar with it. Maybe it's in the US? I recall
                there's a Mount Davis in West Virginia, part of the Appalachian Mountains.
                Let me check its elevation. From what I remember, Mount Davis is around
                4,863 feet (about 1,482 meters).  Now, Grand Colombier. That's in France,
                part of the Alps. I think Grand Colombier's elevation is higher. Let me
                confirm. Grand Colombier is approximately 5,085 feet (around 1,547 meters).
                Wait, no, maybe I'm mixing it up with another peak. Wait, no, actually, I
                might be confusing it with other Alps mountains. Let me double-check. Oh,
                wait, maybe Grand Colombier is higher than Mount Davis. So if Grand
                Colombier is around 1,547 meters and Mount Davis is 1,482 meters, then Mount
                Davis is lower. So the answer should be Mount Davis.
                \\[1em]
                \midrule
                \textbf{Model:} \textsc{Qwen3-8B}
                \\[2pt]
                Dragon's Breath (chili pepper) predicted: $1000000$ actual: $2480000$ \\[2pt]
                Trinidad Scorpion Butch T predicted: $1400000$ actual: $1463700$      \\[2pt]
                \textbf{Trace:} Okay, the user is asking which pepper is milder between Dragon's Breath and
                Trinidad Scorpion Butch T. They want the answer based on the Scoville scale.
                First, I need to recall the Scoville ratings for both. Dragon's Breath is
                known to be extremely hot, around 2.5 million SHU. The Trinidad Scorpion
                Butch T is also very spicy, but I think it's slightly less than Dragon's
                Breath. Wait, let me confirm.   Wait, actually, the Trinidad Scorpion Butch
                T is about 1.4 million SHU, while Dragon's Breath is around 2.5 million. So
                the Trinidad Scorpion Butch T is milder. But wait, I might be mixing up some
                numbers. Let me double-check.   No, I think the Dragon's Breath is higher.
                Wait, some sources say Dragon's Breath is up to 2.5 million, while Butch T
                is around 1.4 million. So Butch T is milder. Therefore, the answer should be
                Trinidad Scorpion Butch T.
                \\[1em]
                \midrule

                \textbf{Model:} \textsc{Qwen3-32B}
                \\[2pt]
                Breg (predicted: $120$km    actual: $49$km)                           \\[2pt]
                Krka (predicted: $76$km     actual: $73$km)                           \\[2pt]
                \textbf{Trace:} Okay, the user is asking which river is shorter between Breg and Krka. Let
                me recall what I know about these rivers.  First, I need to remember the
                lengths of both rivers. Breg River is located in Croatia, I think. It flows
                into the Adriatic Sea. From what I remember, the Krka River is also in
                Croatia and is known for its waterfalls, like the Krka National Park.
                Wait, I'm not entirely sure about their lengths. Let me try to recall. The
                Krka River is a bit longer than Breg. I think Krka is around 70 kilometers
                long, while Breg is shorter, maybe around 30 kilometers? Or is it the other
                way around? Hmm. Wait, no, I might be mixing them up. Let me check my memory
                again.   The Krka River is famous for its series of waterfalls and is a
                popular tourist spot. If it's longer, that would make sense because longer
                rivers can have more features like waterfalls. Breg River is a smaller river
                that flows through the town of Breg.   Wait, actually, I might have confused
                them. Maybe Breg is longer? No, I think Krka is longer. Let me confirm.
                Another way to think: Krka is part of the larger water system, maybe
                connected to the Adriatic Sea through a longer course. Breg might be a
                tributary or a shorter river.   I think the correct answer is that Breg
                River is shorter than Krka River. So the answer should be Breg.
                \\[2pt]
                \bottomrule
        \end{tabularx}
        \caption{Exemplary reasoning traces}
        \label{tab:cot-traces}
\end{table*}

\end{document}